\documentclass[pmlr]{jmlr}

\usepackage{longtable}

\usepackage{booktabs}
\usepackage[load-configurations=version-1]{siunitx} 
\usepackage{multibib}
\newcites{sup}{Supplementary References}
\usepackage{microtype,booktabs,xcolor}

\definecolor{Gray}{gray}{0.85}
\usepackage{algpseudocode}

\usepackage{bm}
\usepackage{bbm}
\usepackage{multirow}
\usepackage{calrsfs}
\usepackage{nicefrac}
\usepackage[symbol]{footmisc}
\usepackage{cancel} 
\renewcommand{\thefootnote}{\fnsymbol{footnote}}
\usepackage{float}
\usepackage[T1]{fontenc}

\usepackage{colortbl}
\usepackage{tcolorbox}
\newtcolorbox{mybox}{colback=white,colframe=black, boxsep=1pt}
\usepackage{caption}

\newcommand{\model}{\textsc{DCM}}

\makeatletter
\def\set@curr@file#1{\def\@curr@file{#1}} 
\makeatother

\theorembodyfont{\upshape}
\theoremheaderfont{\scshape}
\theorempostheader{:}
\theoremsep{\newline}

\jmlrvolume{126}
\jmlryear{2021}
\jmlrworkshop{Machine Learning for Healthcare}

\usepackage{amsmath,amsfonts,bm}

\def\eqref#1{equation~\ref{#1}}

\def\1{\bm{1}}

\def\vx{{\bm{x}}}

\def\mS{{\bm{S}}}

\DeclareMathAlphabet{\mathsfit}{\encodingdefault}{\sfdefault}{m}{sl}
\SetMathAlphabet{\mathsfit}{bold}{\encodingdefault}{\sfdefault}{bx}{n}

\title[Deep Cox Mixtures]{Deep Cox Mixtures for Survival Regression}

\author{\Name{Chirag Nagpal$^{1,2}$, Steve Yadlowsky$^{1}$, Negar Rostamzadeh$^{1}$ and Katherine Heller$^{1}$}
      \Email{chiragn@cs.cmu.edu}\\ 
      \addr $^{1}$Brain Team, Google Research\\
      $^{2}$Auton Lab, Carnegie Mellon University
      }

\begin{document}

\maketitle

\begin{abstract}
Survival analysis is a challenging variation of regression modeling because of the presence of censoring, where the outcome measurement is only partially known, due to, for example, loss to follow up. Such problems come up frequently in medical applications, making survival analysis a key endeavor in biostatistics and machine learning for healthcare, with Cox regression models being amongst the most commonly employed models. We describe a new approach for survival analysis regression models, based on learning mixtures of Cox regressions to model individual survival distributions. We propose an approximation to the Expectation Maximization algorithm for this model that does hard assignments to mixture groups to make optimization efficient. In each group assignment, we fit the hazard ratios within each group using deep neural networks, and the baseline hazard for each mixture component non-parametrically.

We perform experiments on multiple real world datasets, and look at the mortality rates of patients across ethnicity and gender. We emphasize the importance of calibration in healthcare settings and demonstrate that our approach outperforms classical and modern survival analysis baselines, both in terms of discriminative performance and calibration, with large gains in performance on the minority demographics.
\end{abstract}

\section{Introduction}
The importance of survival analysis models in medical applications cannot be overstated. These models support physicians and epidemiologists in clinical decision making based on data-driven evidence about patients' likelihood of survival characteristics based on biological measurements and demographic information about the patients. In this paper, we focus on estimating the patient's risk of an event $T$ of interest, specifically the conditional survival curve, $\mathbb{P}(T>t|X)$. Typically events include death, or the presence or progression of a health condition.

The one frequent challenge with estimating the survival curve is that outcomes are typically censored, meaning that the outcome is unknown for some patients due to lack of follow up or independent competing events. Luckily, censoring is relatively straightforward to deal with in certain commonly used survival analysis models that make the proportional hazards assumption, such as the Cox regression model, or Faraggi-Simon deep neural network model. Unfortunately, in many important cases, the proportional hazards assumption does not hold, leading to poor calibration of patients' estimated survival curve, even if the model can rank patients well.

In fact, many recent deep learning approaches demonstrate significant improvement in ranking patients' survival according to discriminative measures such as the concordance index ($C$-index). However, the $C$-index measures pairwise ranking ability and disregards the absolute value of the actual estimated risk score akin to metrics of evaluating binary classification like the Receiver Operation Characteristic.

In this paper we propose, `\textbf{Deep Cox Mixtures}' for survival analysis, which generalizes the proportional hazards assumption via a mixture model, by assuming that there are latent groups and within each, the proportional hazards assumption holds. Our approach allows the hazard ratio  in each latent group, as  well as the latent group membership, to be flexibly modeled by a deep neural network, allowing us to take advantage of  the recent improvements in neural network modeling of patient data.

In our experiments, we show that the added flexibility of this mixture of proportional hazards models allows us to improve the calibration of the estimated conditional survival curves, while maintaining excellent discriminative performance; that is, without requiring a performance trade-off. We find that the largest improvements to calibration occur among minority groups, and emphasize the need for evaluating performance on such groups, which can often go unnoticed on dataset-wide performance statistics. Our model is implemented in \texttt{tensorflow} and source code of our experiments is open source and publicly available at \url{{https://github.com/chiragnagpal/deep_cox_mixtures}}.\\

\noindent \textbf{Technical Significance} The proposed Deep Cox Mixtures model is not restricted by the strong assumption of proportional hazards. By allowing the model to flexibly choose these latent groups, we can build a more expressive survival analysis model. However inference is challenging owing to the fact that the Cox model involves learning the baseline survival distributions non-parametrically. We develop an approximate Monte Carlo Expectation-Maximization (EM) learning algorithm to estimate the latent groups and parameters of conditional survival curves within each group. To make the learning algorithm tractable, we propose to approximate the Maximization step (M-step) by hard assignment of each patient to a latent group, and approximate the baseline survival curves in the Expectation step (E-step) with spline estimation\\

\noindent \textbf{Clinical Relevance} 
Survival analysis methods can help healthcare practitioners determine risk, triage and support clinical decision making. However studies show systemic miss-estimation of the prognosis and risk by statistical approaches on some demographics, can lead to wrong and harmful decision making~\citep{vyas2020hidden}. One example is the 2013 ACC/AHA\footnote{American College of Cardiology/American Heart Association} Pooled Cohort Equations (PCE) to asses cardio-vascular risk \citep{stone20142013}. \citeauthor{yadlowsky2018clinical} demonstrate that 2013 PCEs overestimate the risk for approximately 11.8 million U.S. adults and this overestimation is especially prominent amongst the black population. In this study, we consider improving calibration across minority demographics as a step towards making more equitable models. In particular we reduce the miss-estimation in underrepresented demographics; their calibration and discriminatory performance upon classical and modern survival analysis baselines. 

\section{Related Work}
Recent progress in deep learning has also sparked interest in the survival analysis community. Recent thrusts in survival analysis have involved deep learning based Cox models \citep{katzman2018deepsurv} like the original Faraggi-Simon network \citep{faraggi1995neural}. More recent papers have explored the use of Discrete time models \citep{lee2018deephit}, recurrent neural architectures \citep{lee2019dynamic} as well as fully parametric methods \citep{nagpal2020deep} for modelling survival outcomes in the presence of censoring. More involved techniques have involved the use of ensembles with black box optimization, auto encoding variational bayes \citep{chapfuwa2020survival, xiu2020variational}, as well as adversarial methods \citep{chapfuwa2018adversarial} to estimate survival outcomes. 

Attempts to learn a mixture of Cox models \citep{nagpal2019nonlinear, rosen1999mixtures}  have focused primarily on learning a mixture of log-linear parametric components for the hazard ratio in the partial log-likelihood. These approaches are still subject to the strong assumptions of proportional hazards.  Towards the best of our knowledge, our approach is the first attempt at learning a Cox mixture model using the full likelihood and jointly estimating both the parametric relative hazard and the baseline hazard functions. Close lines of work to ours include \cite{chapfuwa2020survival, ranganath2016deep}, where the authors propose the latent space to be a mixture distribution and sample the outcome event time from a parametric decoder. Our approach differs from this as we do not need to make any strong parametric assumptions on the event outcome times.

There has also been an interest in learning survival models on time-series and temporal data \citep{lee2019dynamic}. In this paper we restrict our approach to the case with static feature snapshots, although since our approach involves representation learning using neural networks, it can be easily extended to these settings with appropriate choice of recurrent neural networks.

Poor calibration of deep learning methods has been explored recently in machine learning literature \citep{guo2017calibration, nixon2019measuring}. Poor calibration of Deep Learning models in areas like Natural Language Processing \citep{nguyen2015posterior} and Computer Vision has also been demonstrated. Existing lines of research to improve calibration have involved post processing techniques like Platt Scaling, Bayesian and ensemble methods as well as IPM penalties \citep {kumar2018trainable} to improve model calibration. Calibration in the specific case of survival models has been an active area of research as well.  \cite{lee2019temporal} proposed an ensemble of multiple survival analysis models weighted using Black-Box Bayesian optimization for better calibration. This makes for an interesting modelling approach but practical application is challenging due to computational complexity.

Literature in algorithmic fairness has proposed calibration over subgroups as a measure of algorithm fairness \citep{kleinberg2016inherent, chouldechova2017fair, pleiss2017fairness}. In these works, calibration is typically referred to as `sufficiency' or `matching conditional frequencies' \citep{hardt2016equality} and evaluated using reliability diagrams. We stress that as opposed to scenarios where an algorithm is employed to determine the assignment to a service, in healthcare we are typically interested in estimating risk. In as much errors on both sides (under and over estimation) of the risk are potentially unfair making calibration a well suited metric for fairness evaluation.\footnote[1]{In healthcare, it is typically ethical to include demographic information like race and gender when estimating outcomes. If there are strong reasons to believe that such information does not cause the outcome, other definitions of algorithmic fairness might be more valid.}

Survival analysis scenarios are also prone to censoring, making estimation of the Expected Calibration Error challenging. Methods involving evaluation for calibration in the presence of censoring have involved simple histogram based binning methods followed by Kaplan-Meier or IPCW estimation of the Survival probability within each bin. More involved recent methods involve non parametric methods like regression splines \citep{austin2020graphical} and kernel methods \citep{yadlowsky2019calibration}. In this tradition, we shine the light on the calibration of models in our empirical evaluations, emphasizing the calibration within minority groups, in particular. We find that without sacrificing discriminative performance, the added flexibility of our mixture model improves calibration, overall and especially in minority groups.

\section{The Deep Cox Mixture Model}

\begin{figure*}[!ht]
    \centering
    \includegraphics[width=\textwidth, trim={0cm 18.5cm 2cm 1.5cm}, clip]{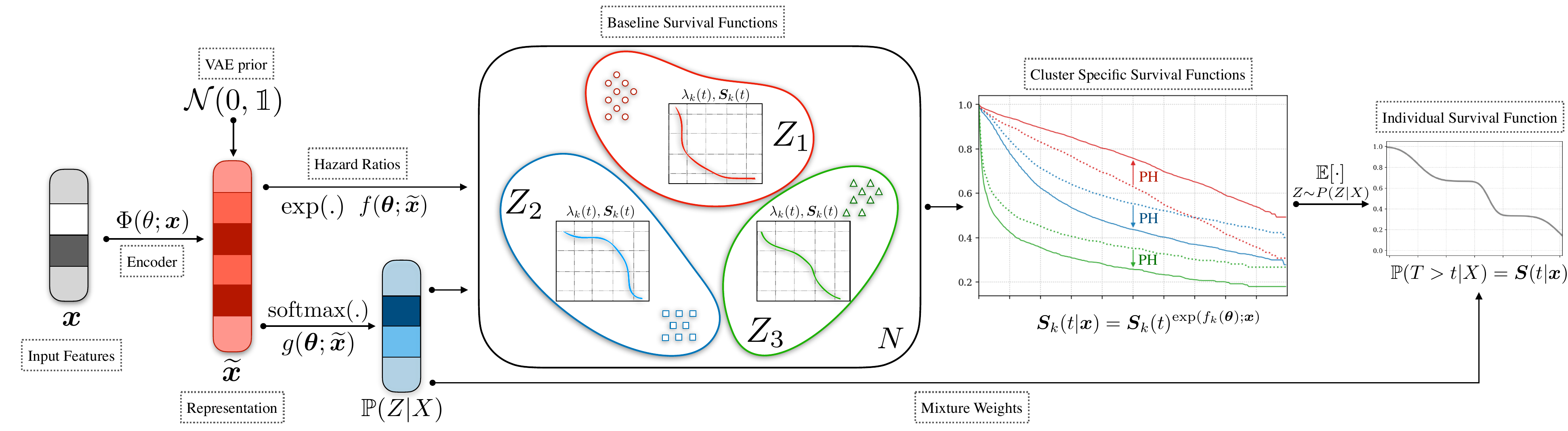}
    \caption{\textbf{Deep Cox Mixtures}: Representation of the individual covariates $\bm{x}$ are generated using an encoding neural network. The output representation $\widetilde{\bm x}$ then interacts with linear functions $f$ and $g$ that determine the proportional hazards within each cluster $Z\in \{ 1, 2, ... K\}$ and the mixing weights $\mathbb{P}(Z|X)$ respectively. For each cluster, baseline survival rates $\bm{S}_k(t)$ are estimated non-parametrically. The final individual survival curve $S(t|\bm x)$ is an average over the cluster specific individual survival curves weighted by the mixing probabilities $\mathbb{P}(Z| X= \vx)$.    }
    \label{fig:dcm}
\end{figure*}

\subsection{Notation}
We consider a dataset of right censored observations $\mathcal{D} = \{ (\vx_i, \delta_i, u_i ) \}_{i=1}^{N}$ of three tuples, where $\vx_i$ are the covariates of an individual $i$, $\delta_i$ is an indicator of whether an event occured or not and $u_i$ is either the time of event or censoring as indicated by $\delta_i$. 

We consider a maximum likelihood (MLE) based approach to learning $S(t | x) = \mathbb{P}(T > t | X=x)$ from the data. Recall that the survival distribution $S(t | x)$ is isomorphic to the cumulative hazard function $\bm\Lambda(t | x)$, and under continuity, this is equivalent to the hazard function $\bm\lambda(t | x)$. As a result, we will refer them in the parameters of the likelihood interchangeably. \citet{lin2007breslow} shows that the likelihood of the observed data $\mathcal{D}$ is, up to constant factors,
\begin{equation}
\label{eq:full-likelihood}
\mathcal{L}(\mathbf{\Lambda}) = \prod_{i=1}^{|\mathcal{D}|} \left( \bm\lambda(u_i | \vx_i) \right)^{\delta_i} \mS(u_i | \vx_i).
\end{equation}
In the following sections, we show how plugging in specific functional forms for $S(t | x)$ allows us to derive survival function estimators.

\subsection{MLE for the standard Cox PH model}

The key idea behind the Cox model is to assume that the conditional hazard of an individual, is $\bm\lambda(t|x) = \bm\lambda_0(t) \exp \big( f(\bm\theta, x) \big)$, where $f$ is typically a linear function. Under the Cox model, the full likelihood as in \eqref{eq:full-likelihood} is
\begin{align}
\mathcal{L}(\bm{\theta},  \mathbf{\Lambda}_0 ) &= \prod_{i=1}^{|\mathcal{D}|} \bigg( \bm\lambda_0(u_i) \exp \big( f(\theta, \vx_i) \big) \bigg)^{\delta_i} \mS_0(u_i) ^{\exp \big(f(\bm{\theta}; \vx_i ) \big)  } 
\end{align}

\citet{cox1972regression} and the discussion of his paper by \citet{breslow1972contribution}, suggest deriving a maximum likelihood estimate of $\bm{\theta}$ by maximizing the partial likelihood, $\mathcal{PL}(\bm\theta)$ defined below, and using the following estimator of the baseline survival function $\mathbf{\Lambda}_0(\cdot)$,
\begin{align}
\label{eqn:pll}
\mathcal{PL}(\bm\theta) =   \prod_{i: \delta_i=1} \frac{ \exp \big( f(\bm\theta; \vx_i) \big)}{\sum\limits_{j\in \mathcal{R}(t_i) } \exp \big( f(\bm\theta; \vx_j) \big)}, \quad \widehat{\mathbf{\Lambda}}_0(t) = \sum\limits_{i:t_i<t}\frac{1}{\sum \limits_ {j\in \mathcal{R}(t_i) }  \exp \big(f(\widehat{\bm{\theta}}; \vx_j ) \big)},
\end{align}
where $\mathcal{R}(t_i)$ is the `risk set' -- the set of individuals that survived beyond time $t_i$.

\subsection{Proposed Model}

In the case of \model{} we propose an extension to the Cox model, modeling an individual's survival function using a finite mixture of $K$ Cox models, with the assignment of an individual $i$ to each latent group mediated by a gating function $g(.)$ The full likelihood for this model is 
\begin{equation*}
\mathcal{L}(\bm\theta, \mathbf{\Lambda}_k) = \prod_{i=1}^{|\mathcal{D}|} \int_Z \left( \bm\lambda(u_i | \vx_i) \right)^{\delta_i} \mS_k(u_i | \vx_i)\mathbb{P}(Z=k|\vx_i).
\end{equation*}
\vspace{-1em}
\begin{align}
\nonumber \text{where, } \bm \lambda(u_i | \vx_i) = \bm\lambda_k(u_i) \exp \big( f_k(\bm\theta, \vx_i) \big), \quad \mS_k(u_i | \vx_i) = \mS_k(u_i)^{\exp \big(f_k(\bm{\theta}; \vx_i ) \big )}\\
\label{eq:model-likelihood} \text{and, } \mathbb{P} (Z=k|X=\vx_i) = \mathrm{softmax}\big( g(\bm\theta; \vx_i )\big)
\end{align}

\noindent \textbf{Architecture:} We allow the model to learn representations for the covariates $\vx_i$ by passing them through a encoding neural network, $\Phi(.):\mathbb{R}^d \to \mathbb{R}^h $. This representation then interacts with linear functions $f$ and $g$ defined on $\mathbb{R}^h \to \mathbb{R}^k$; that determine the log hazard ratios and the mixture weights respectively. The set of parameters for the encoder $\Phi$ and the linear functions $f$ and $g$ are jointly notated as $\bm \theta$.  We experiment with a simple feed forward MLP and a variational auto-encoder for $\Phi(.)$ The parameters of the MLP and the VAE are learnt jointly during learning. For the VAE variant the encoder and the decoder architecture is kept the same. We also experiment with a variant that doesn't use representation learning and thus the functions $f$ and $g$ are linear and restricted to operate on the original features $\bm x$. Figure \ref{fig:dcm} provides a schematic description of our approach.

\subsection{Learning}

\begin{minipage}{0.4\textwidth}
Notice that under the model in Eq. \ref{eq:model-likelihood}, the corresponding partial likelihood is not independent of $\bm \lambda(.)$, the hazard rate. We hence cannot directly optimize the partial likelihood to perform parameter learning. This inference complexity is outlined in Appendix \ref{apx:na}. Since our model requires inference over the latent assignments $Z$ for learning the Expectation Maximization \citep{dempster1977maximum} algorithm is a natural approach to perform inference. The major challenge to applying exact EM lies in the fact that under the our model requires a summation over all possible combinations of latent assignments and which is intractable to compute. We propose an approximate, Monte Carlo EM  
\end{minipage}%
\hfill
\begin{minipage}{0.58\textwidth}
\begin{mybox}
\begin{algorithm}[H]
 \caption{\textbf{Learning for \model{} }}
  \label{alg:algo}
  \hrule
\SetAlgoLined
  \SetKwInOut{Input}{Input}
  \SetKwInOut{return}{Return}
  \Input{Training set, $\mathcal{D} = \{ (\vx_{i}, t_i, \delta_i)_{i=1}^{N}  \}$; \\ batches, $B$;  }
\hrule
 \While{\texttt{<not converged>}}{
  \For{$b \in \{1, 2, ...,  B \} $ }{
    $\mathcal{D}_b \gets $\textbf{\texttt{sampleMiniBatch}} ({$\mathcal{D}$})  
    
    $\{ {\gamma_i} \}_{i=1}^{B} \gets $ \textbf{{E-Step}}( $\bm{\theta}, \{\widetilde \mS_{k} \}_{i=1}^{K}$ ) \;
    
    $\{ {{\zeta}_i} \}_{i=1}^{B} \sim $ \text{Categorical}(${\gamma} )$ \;
    
    $\bm{\theta} \gets $ \textbf{M-Step}($\bm{\theta}, \{ {\zeta_i, \gamma_i} \}_{i=1}^{B} $)\;
    
  \For{$k \in \{1, 2, ..., K \}$  }{
    $\widehat \mS_{k} \gets \textbf{\texttt{breslow}}(\bm{\theta}, \{(t_i, \delta_i)\}_{i=1; \zeta_i = k}^{|\mathcal{D}|})$ \;
    
    $\widetilde \mS_{k} \gets $\textbf{\texttt{splineInterpolate}}($\widehat{\mS_{k}}$) \;
  }
 }
}
\hrule \vspace{.2em}
\return{learnt parameters, $\bm{\theta}$;\\ baseline survival splines $\{\widetilde{\mS_{k}} \}_{i=1}^{K}  $}
\end{algorithm}
\end{mybox}
\end{minipage}

\noindent  algorithm \citep{wei1990monte, song2016learning}  involving the drawing of posterior samples to learn the parameters, $\bm{\theta}$ and the baseline survival functions $\{ \mS_{k}(.)\}_{i=1}^{K}$.\\

\noindent \textbf{E-Step}: Involves estimating the posteriors of $Z$, $\gamma_i \propto \mathbb{P}(T=t|X, Z)^{\delta_i}\mathbb{P}(T>t|X, Z)^{1-\delta_i}$. The Breslow estimator only gives us the estimates of the survival rates, thus computing the posterior counts, $h_i \propto \mathbb{P}(T=t_i|Z, X ) $ for the uncensored instances challenging. We mitigate this by interpolating the Baseline Survival Rate for each latent group, $\mS_k(.)$ using a polynomial spline. Equation \ref{eqn:uncens_prob} provides the interpolated event probability estimates. (Appendix {\ref{apx:model_splines}} describes this in detail.)
\begin{align}
 &\widehat{\mathbb{P}}(T>t|X=\vx_i, Z=k) = \widetilde{\mS}_k(t) ^{\exp \big(f_k(\bm{\theta}; \vx_i ) \big)  }\text{ and, } \nonumber\\
 \label{eqn:uncens_prob}&\widehat{\mathbb{P}}(T=t|X=\vx_i, Z=k) = \nonumber\\ 
 &\hspace{6em} - \exp \big(f_k(\bm{\theta}; \vx_i ) \big)   \frac{ \widehat{\mathbb{P}}(T>t|\vx_i, Z=k))}{\widetilde{\mS}_k(t)} \frac{ \partial }{\partial t}  \widetilde{\mS}_k(t) 
\end{align}
Here, $\widetilde{\mS}_k(t) $ is the baseline survival rate interpolated with a polynomial spline.\\


\noindent \textbf{M-Step}: Once the posterior counts $\gamma_i$ are obtained, the M-Step involves learning maximizing the corresponding $Q(.)$ function given as 
\begin{align}
 \nonumber Q(\theta) = \sum \limits_{i=1}^{|\mathcal{D}|}  \sum \limits_{k}  \gamma^{k}_i \ln   \mathbb{P}(Z|X) &+ \gamma^{k}_i \ln \mathbb{P}(t|Z,X);\\  \text{where, }\gamma_i  &\propto \mathbb{P}(T|X,Z)
\end{align}

Notice that the $\gamma^k_i$ are soft counts ($\gamma_i \in [0,1]$) making parameter inference for the term $\mathbb{P}(T|Z, X)$ intractable. Motivated from Monte-Carlo EM methods  We instead sample hard posterior counts $\zeta_i \sim \mathrm{Categorical}(\gamma_i)$.

We replace this with hard posterior counts for the second term, $\ln \mathbb{P}(t|Z,X)$
\begin{align}
  \nonumber \overline{Q}(\theta) = \sum \limits_{i=1}^{|\mathcal{D}|}  \sum \limits_{k}  \gamma^{k}_i \ln   \mathbb{P}(Z|X) &+ \zeta^{k}_i \ln \mathbb{P}(t|Z,X);\\  \text{where, }\zeta_i &\sim \mathrm{Categorical}(\gamma_i )
\end{align}
Note that $\mathbb{E}[\overline{Q}(\cdot)] = Q(\cdot)$. Thus, $\overline{Q}(\cdot)$ is an unbiased estimate of the exact $Q(\cdot)$

The first term in $\overline{Q}(\cdot)$ can be optimized using gradient based approaches. The second term can be re-written as a sum over $k$ latent groups variables.
\begin{align}
\nonumber  \overline{Q}(\theta) &= \sum \limits_{i=1}^{|\mathcal{D}|}  \sum \limits_{k}  \gamma^{k}_i \ln   \mathbb{P}(Z|X) + \mathbbm{1}\{\zeta_i =k\} \ln \mathbb{P}(t|Z,X)\\
 \nonumber &= \sum \limits_{i=1}^{|\mathcal{D}|}  \sum \limits_{k}  \gamma^{k}_i \ln   \mathbb{P}(Z|X) + \sum \limits_{i=1}^{|\mathcal{D}|}  \sum \limits_{k} \mathbbm{1}\{\zeta_i =k\} \ln \mathbb{P}(t|Z,X)\\
  &= \sum \limits_{i=1}^{|\mathcal{D}|}  \sum \limits_{k}  \gamma^{k}_i \ln   \mathbb{P}(Z|X) + \sum \limits_{k} \sum \limits_{i=1}^{|\mathcal{D}_k|} \ln \mathbb{P}(t|Z,X)\\ & \qquad\qquad\qquad \text{(Here, $\mathcal{D}_k$ is the set of all $\mathcal{D}$ with $\zeta_i = k$)} \nonumber
\end{align}

Now using the fact that the Proportional Hazards assumption holds within each group $\mathcal{D}_k$ we arrive at the form of the $Q(\cdot)$ that we optimize in each minibatch as 
\begin{align*}
  \widehat{Q}(\bm \theta) = \sum_i^{|\mathcal{D}_b|} \sum_k \gamma^k_i \ln  {\mathrm{softmax}\big( g(\bm\theta; \vx_i )\big)} + \sum_k \ln  \mathcal{PL}(\mathcal{D}^{k}_b; \bm\theta)
\end{align*}

Here, $\mathcal{D}_b^{k}$ is the subset of all individuals that have $\zeta_i = k$ within the minibatch $b$ and $\mathcal{PL}(.)$ is the partial likelihood as defined in Equation \ref{eqn:pll}. Thus, the use of hard counts $\zeta$ effectively reduces the problem to learning $K$ separate Cox models allowing us to maximize the partial likelihood independently within each $k\in K$.


The parameters of the encoder are also updated during the \textbf{M-Step} by adding the loss corresponding to the VAE. Altogether the loss function for optimization is 
\begin{align}
\text{Loss}(\bm\theta; \mathcal{D}_b) = \widehat{Q}(\bm\theta ; \mathcal{D}_b) + \alpha\cdot\textrm{VAE-Loss}( \bm \theta ; \mathcal{D}_b)
\end{align}

Here, the $\textrm{VAE-Loss}$ is the Evidence Lower Bound for the VAE with representations drawn from a zero mean and identity covariance gaussian prior as in \citesup{kingma2013auto}.

Algorithm \ref{alg:algo} describes the learning procedure for \model{}. We sample minibatches $\mathcal{D}_b$ from the data $\mathcal{D}$ and compute the soft and hard posterior counts, $\{ \gamma_i$, $\zeta_i \}_{i\in\mathcal{D}_b}$ for each batch. This is followed by the \textbf{M-Step} involving a gradient update the parameter set $\bm\theta$. Finally, we update the Baseline Survival Splines,  $\widetilde \mS_k$ computed using the Breslow's estimator (Eq. \ref{eqn:pll}) for each cluster. Note that the Breslow's estimator is computed over the full batch, $\mathcal{D}$. This is computed analytically, does not involve gradient computation and so is not expensive.

\subsection{Inference}

Following Equation \ref{eq:model-likelihood}, at test time the estimated risk of an individual at time $t$ is given as 
\begin{align}
   \nonumber \widehat{\mathbb{P}}&(T>t|X=\vx_i) = \mathbb{E}_{Z\sim \widehat{\mathbb{P}}(Z|X)}[ \widehat {\mathbb{P}}(T|X=\vx_i, Z)] \\ 
    &= \sum_k \widetilde \mS_k(t)^ { \exp \big(f(\bm{\theta}; \vx_i ) \big)  } \times \mathrm{softmax}_k\big(g(\bm{\theta}; x_i)\big)
\end{align}

\section{Experiments}

\begin{table*}[!ht]
    \centering
    \caption{Summary statistics for the datasets used in the experiments.}
    \centerline{\resizebox{\textwidth}{!}{\begin{tabular}{l|r|r|r|r|r|r|r}    
    \toprule \midrule
    \multirow{2}{*}{Dataset}& \multirow{2}{*}{$N$} &\multirow{2}{*}{$d$} & \multirow{2}{*}{Censoring ($\%$)} &\multirow{2}{*}{Minority Class (\%)}& \multicolumn{3}{c}{Event Quantiles}  \\ \cline{6-8}
    & & & & &$t=25$th&$t=50$th&$t=75$th  \\ \midrule
         \textsc{SUPPORT} &9,105& 44& 31.89\%&Non-White (21.02\%) &14& 58& 252  \\
         \textsc{FLCHAIN} &6,524 & 8 &69.93\% &Female (44.94\%)& 903.25& 2085&   3246\\
         \textsc{SEER} &55,993&168&72.82\%&Non-White (23.77\%)& 25&  55& 108 \\ \bottomrule
    \end{tabular}}}
    \label{tab:datasets}
\end{table*}

In this section we describe the datasets, the survival analysis tasks and baselines we compare \model{} against. We also describe the corresponding metrics we employ for evaluation.

\subsection{Datasets}

We experiment with the following real world, publicly available survival analysis datasets:\\

\noindent \textbf{\textsc{FLCHAIN}} (Assay of Serum Free Light Chain):  This is a public dataset introduced by \cite{dispenzieri2012use} aiming to study the relationship between serum free light chain and mortality. It includes covariates like age, gender, serum creatinine and presence of monoclonal gammapothy. We removed all the individuals with missing covariates and experiment with the remaining subset of 6,524 individuals. Out of this subset $45\%$ of the participants were coded as female and are considered as `minority' in our experiments.\\

\noindent \textbf{\textsc{SUPPORT}} (Study to understand prognoses and preferences for outcomes and risks of treatments \citep{connors1995controlled}: Dataset from study instituted to understand patient survival for 9,105 terminally ill patients on life support. The median survival time for the patients in the study was 58 days. Out of the $9,105$ patients a majority 79\% were coded as `White', while the rest were coded as `Black', `Hispanic' and `Asian'.\\

\noindent \textbf{\textsc{SEER}} (Surveillance, Epidemiology and End Results Study)\footnote{\url{https://seer.cancer.gov/}} : This dataset from \cite{nci2019seer} consists of survival characteristics of oncology patients taken from cancer registries covering about one-third of the US Population. For our study we consider a cohort of patients over a 15 year period from 1992-2007 diagnosed with breast cancer with a median survival time of 55 months. A majority (76\%) of the patients were coded as `White' and the rest were other minorities consisting of `Blacks', `American Indians', `Asians', etc.\footnote{\textsc{SEER} has a very intricate coding pattern vis-a-vis race. Refer to \url{https://seer.cancer.gov/tools/codingmanuals/race_code_pages.pdf} for details. }

\begin{figure}[!tbp]
    \centering
    $\qquad\qquad\qquad\qquad\quad$ SEER \hfill SUPPORT $\quad\qquad\qquad\qquad$
    \includegraphics[width=0.5\linewidth]{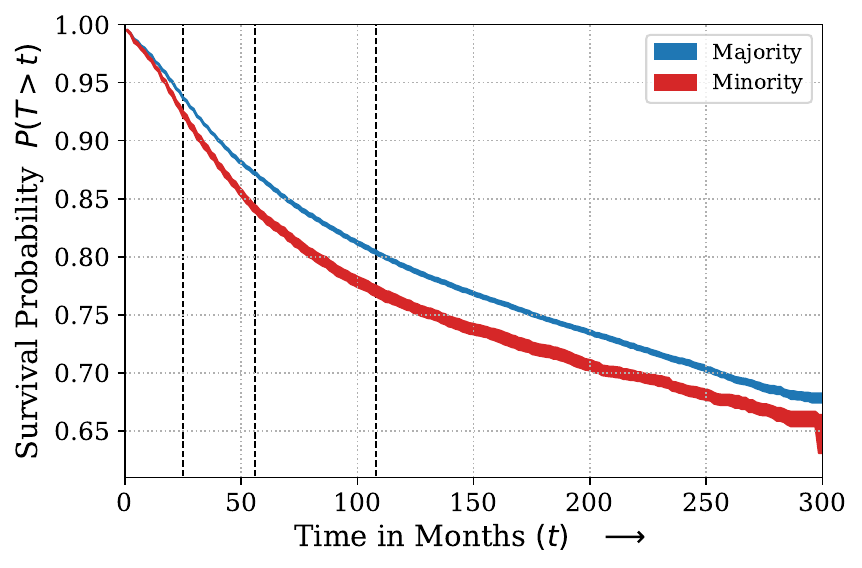}%
    \includegraphics[width=0.5\linewidth]{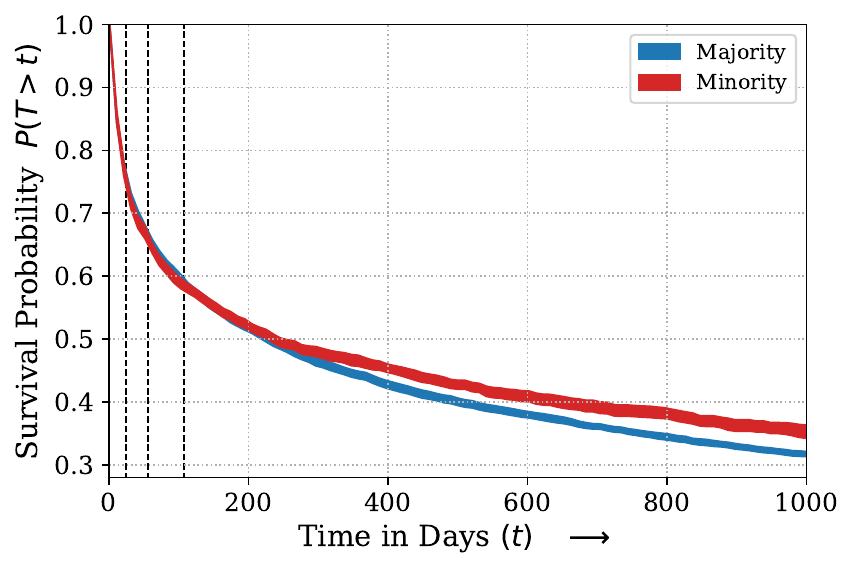}

    \caption{Base survival rates for the majority (White) vs. the other demographics in the \textsc{SEER} dataset estimated with a Kaplan-Meier estimator. Notice that the baseline survival rates differ across groups. Dashed lines respresent the $25^{\text{th}}, 50^{\text{th}}$ and $75^{\text{th}}$ event quantiles.}
    \label{fig:population_survival} 
\end{figure}

Our choice of datasets encompass varying ranges of dimensionality of covariates, levels of censoring and size vis-a-vis the minority demographics. Table \ref{tab:datasets} describes some summary statistics of the considered datasets. Figure \ref{fig:population_survival} compares the baseline survival rates for the majority and minorities in the SEER and SUPPORT dataset. Notice that base survival rates across demographics can vary considerably over time. 
 
\subsection{Baselines}

We compare the proposed \model{} against the following
baselines.\\


\noindent \textbf{Accelerated Failure Time  (AFT)}: This is an extension of generalized linear models to the survival setting with censored data. The target variable is assumed to follow a Weibull distribution and the shape and scale parameters are modelled as linear functions of the covariates. Parameter learning is performed using Maximum Likelihood Estimation.\\

\noindent \textbf{Deep Survival Machines (DSM)} \citep{nagpal2020deep}: This is another fully parametric approach and improves on the Accelerated Failure Time model by modelling the event time distribution as a fixed size mixture over Weibull or Log-Normal distributions. The individual mixture distributions are themselves parametrized with neural networks allowing to learn complex non-linear representations of the data.\\

\noindent \textbf{Deep Hit (DHT)} \citep{lee2018deephit}: A discrete time model, DeepHit is a popular Neural Network approach that involves discretizing the event outcome space and treating the survival analysis problem as a multiclass classification problem over the discrete intervals. \\


\noindent \textbf{Cox Proportional Hazards (CPH)}: CPH assumes that individuals across the population have constant proportional hazards overtime.\\

\noindent \textbf{Faraggi-Simon Net (FSN)/DeepSurv} \citep{faraggi1995neural,katzman2018deepsurv}: An extension to the CPH model, FSN involves modelling the proportional hazard ratios over the individuals with Deep Neural Networks allowing the ability to learn non linear hazard ratios.\\


\noindent \textbf{Random Survival Forest (RSF)} \citep{ishwaran2008random}: RSF is an extension of Random Forests to the survival settings where risk scores are computed by creating Nelson-Aalen estimators in the splits induced by the Random Forest.\\
\begin{mybox}
\textbf{Note:} In practice we observe that performance of the \textbf{Random Survival Forest} model, especially in terms of calibration is strongly influenced by the choice for the hyper-parameters, \textbf{\texttt{mtry}} (the number of features considered at each split) and \textbf{\texttt{min\_node\_size}} (the minimum number of data samples to continue growing a tree). We thus advise carefully tuning these hyper-parameters while benchmarking \textbf{RSF}.
\end{mybox}

The full set of hyper parameter we perform grid search on is deferred to Appendix \ref{apx:bl_hyper}.

\subsection{Evaluation Metrics}

We compare the performance of \model{} against baselines in terms of both discriminative performance and calibration using the following metrics:\\

\noindent \textbf{Area under ROC Curve} (AUC): Involves treating the survival analysis problem as binary classification at different quantiles of event times and computing the corresponding area under the ROC curve.\\

\noindent\textbf{Time Dependent Concordance Index} ($C^\text{td}$): Concordance Index estimates ranking ability by exhaustively comparing relative risks across all pairs of individuals in the test set. We employ the `Time Dependent' variant of Concordance Index that truncates the pairwise comparisons to the events occurring within a fixed time horizon. 
\begin{align*}
C^{td }(t) = \mathbb{P}\big( \hat{F}(t| \mathbf{x}_i) > \hat{F}(t| \mathbf{x}_j)  | \delta_i=1, T_i<T_j, T_i \leq t \big) 
\end{align*}
\noindent \textbf{Expected $\mathbf{\ell_1}$ Calibration Error }(ECE): The ECE measures the average absolute difference between the observed and expected (according to the risk score) event rates, conditional on the estimated risk score. At time $t$, let the predicted risk score be $R(t) = \widehat{\mathbb{P}}(T>t | X)$. Then, the ECE approximates
 \begin{align*}
\text{ECE}(t) = \mathbb{E} \big[ \big| \mathbb{P}(T > t | R(t)) - R(t)  \big| \big]
\end{align*}
by partitioning the risk scores $R$ into $q$ quantiles $\{[r_j, r_{j+1})\}_{j=1}^q$.\\

\noindent \textbf{Brier Score} (BS): The Brier Score involves computing the Mean Squared Error around the binary forecast of survival at a certain event quantile of interest. Brier Score is a proper scoring rule and can be decomposed into components that measure both discriminative performance and calibration.
\begin{align*}
\text{BS}(t) = \mathbb{E}_{\mathcal{D}}\big[ \big(\mathbbm{1}\{ T > t \} - \widehat{\mathbb{P}}(T>t|X)\big)^2  \big]
\end{align*}
Each of the metrics described above are adjusted for censoring by using standard Thompson-Horvitz style Inverse Propensity of Censoring Weights (IPCW) estimates learnt with a Kaplan-Meier estimator over the censoring times. Details are in Appendix \ref{apx:censoring}.

\subsection{Experimental Protocol}

For the proposed model, \model{} and the baselines we perform 5-fold cross validation. The predictions of each fold at the 25th, 50th and 75th quantiles of event times are collapsed together and bootstrapped in order to generate standard errors. For the proposed model and the baselines we report the mean of the evaluation metric and the bootstrapped\footnote{100 times} standard errors for the model that has the lowest Brier Score amongst all the competing set of hyper parameter choices. For DCM, the set of hyperparameter choices include the number of hidden layers for $\Phi$ tuned from $\mathtt{\{1, 2\}}$, units in each hidden layer selected from $\mathtt{ \{50, 100\}}$, the number of mixture components $K$ which are tuned between $\mathtt{ \{3, 4, 6\}}$ and the discounting factor for the VAE-Loss, $\alpha$ tuned from $\mathtt{\{0, 1\}}$. Optimization is performed using the Adam optimizer \citep{kingma2014adam} in \texttt{tensorflow} with learning rates fixed  $\mathtt{1\times 10^{-3}}$ and mini batch size of $\mathtt{128}$. The Baseline Survival Splines are fixed to be of degree $\mathtt{3}$ and  fit using the \texttt{scipy} python package.

\section{Results}

\begin{figure}[!t]
    \centering
    \centerline{{\scriptsize FLCHAIN}}
    {\hspace{5em} \scriptsize Time Dependent Concordance Index \hfill Expected Calibration Error\hspace{7.5em}\phantom{} }\\
    \includegraphics[width=0.445\linewidth]{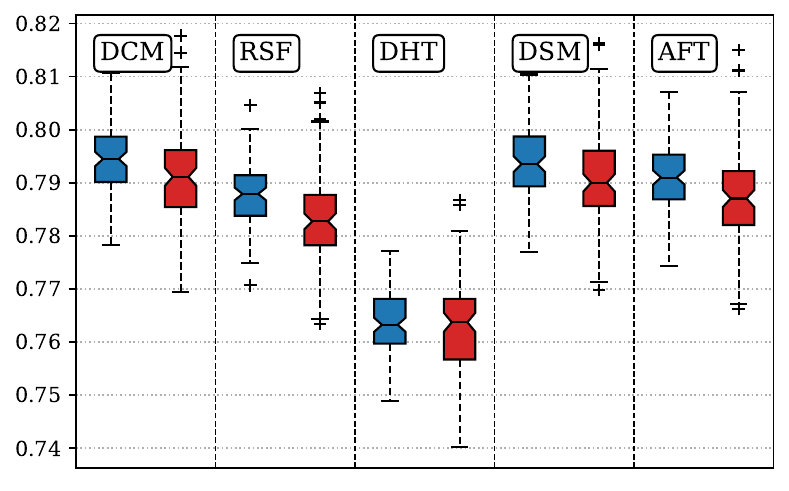}%
    \includegraphics[width=0.445\linewidth]{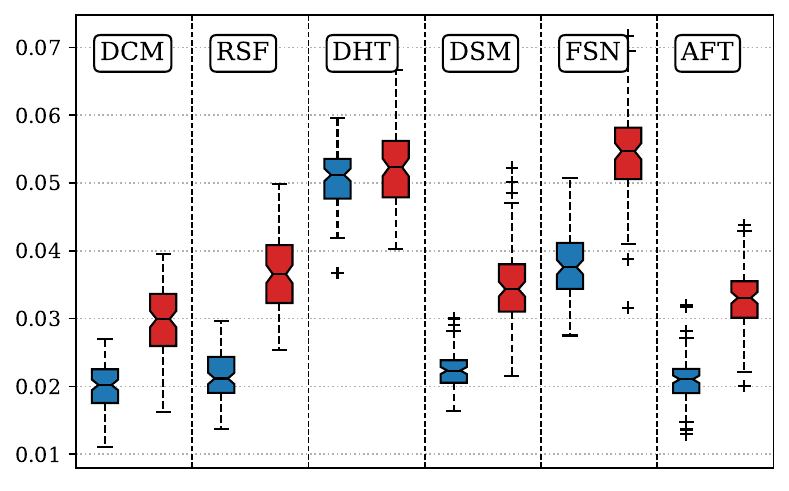}\\
\resizebox{.875\textwidth}{!}{%
   \hspace{1.2em} \begin{tabular}{l|c|c|c|c}
    \toprule \midrule
     \multirow{2}{*}{ \textbf{Model}} & \multicolumn{2}{c|}{\textbf{$C^{\text{td}}$}} & \multicolumn{2}{c|}{\textbf{ECE}}  \\ \cline{2-5} 
& {Population} & {Minority} & {Population} & {Minority}   \\ \hline
CPH& 0.6621 $\pm$ 0.0087 & 0.6737 $\pm$ 0.0124 & 0.0992 $\pm$ 0.0044 & 0.0878 $\pm$ 0.0071  \\
AFT& 0.7911 $\pm$ 0.0060 & 0.7875 $\pm$ 0.0087 & 0.0212 $\pm$ 0.0034 & 0.0329 $\pm$ 0.0046 \\
RSF& 0.7880 $\pm$ 0.0059 & 0.7830 $\pm$ 0.0089 & 0.0215 $\pm$ 0.0037 & 0.0368 $\pm$ 0.0053 \\
FSN& 0.6608 $\pm$ 0.0081 & 0.6212 $\pm$ 0.0131 & 0.0381 $\pm$ 0.0046 & 0.0545 $\pm$ 0.0068  \\
DHT& 0.7636 $\pm$ 0.0059 & 0.7631 $\pm$ 0.0092 & 0.0505 $\pm$ 0.0041 & 0.0525 $\pm$ 0.0056 \\
DSM& 0.7937 $\pm$ 0.0061 & 0.7909 $\pm$ 0.0087 & 0.0223 $\pm$ 0.0029 & 0.0347 $\pm$ 0.0056  \\ \midrule
\rowcolor{Gray}DCM&0.7943 $\pm$ 0.0103 & 0.7911 $\pm$ 0.0091 & 0.0200 $\pm$ 0.0034 &  0.0294 $\pm$ 0.0049  \\ \midrule \bottomrule
\end{tabular}}

\caption{\small $C^{\text{td}}$ (higher means better discrimination) and ECE (lower means better calibration) of proposed approach versus baselines at the $75^{\text{th}}$ event quantile for FLCHAIN.}
\label{fig:flchain_results_75th} 
\end{figure}

\begin{figure}[H]

\centering
\centerline{{\scriptsize SUPPORT}}
    {\hspace{5em} \scriptsize Time Dependent Concordance Index \hfill Expected Calibration Error\hspace{7.5em}\phantom{} }\\
    \includegraphics[width=0.445\linewidth]{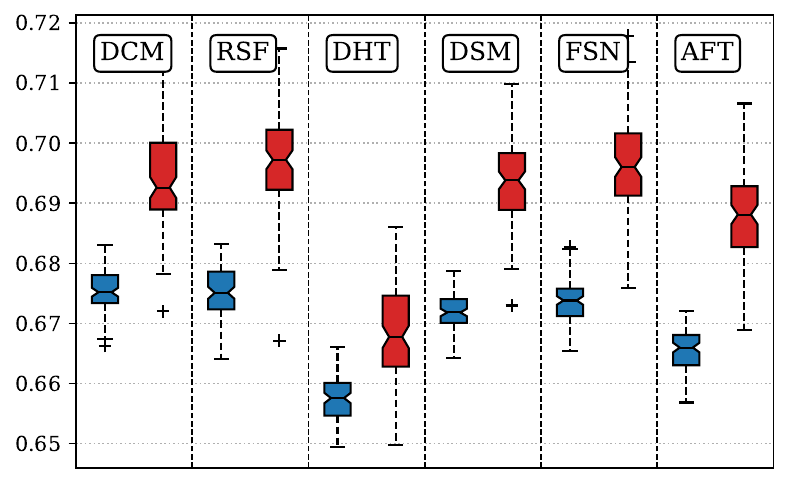}%
    \includegraphics[width=0.445\linewidth]{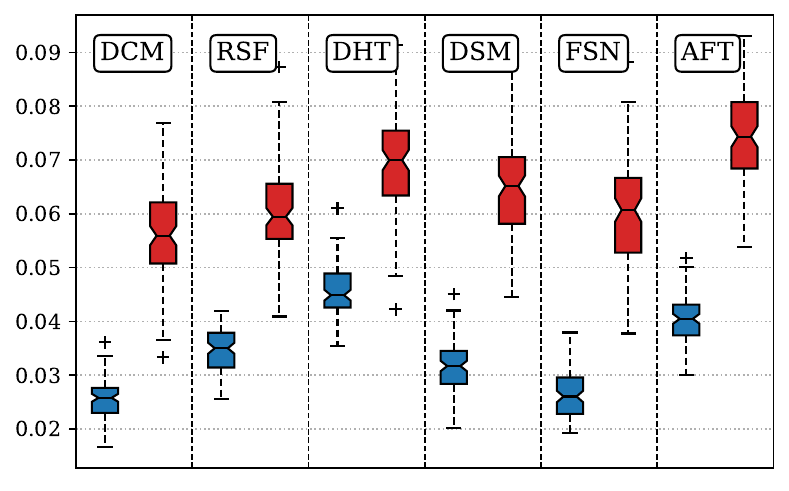}\\
    
    \resizebox{.875\textwidth}{!}{%
    \hspace{1.2em} \begin{tabular}{l|c|c|c|c}
    \toprule \midrule
     \multirow{2}{*}{ \textbf{Model}} & \multicolumn{2}{c|}{\textbf{$C^{\text{td}}$}} & \multicolumn{2}{c|}{\textbf{ECE}}  \\ \cline{2-5} 
& {Population} & {Minority} & {Population} & {Minority}   \\ \hline
CPH&0.6686 $\pm$ 0.0034 & 0.6905 $\pm$ 0.0078 & 0.0310 $\pm$ 0.0041 & 0.0685 $\pm$ 0.0079 \\
AFT&0.6657 $\pm$ 0.0034 & 0.6883 $\pm$ 0.0078 & 0.0402 $\pm$ 0.0046 & 0.0741 $\pm$ 0.0085 \\
RSF&0.6751 $\pm$ 0.0040 & 0.6974 $\pm$ 0.0084 & 0.0348 $\pm$ 0.0041 & 0.0603 $\pm$ 0.0080  \\
FSN&0.6736 $\pm$ 0.0037 & 0.6961 $\pm$ 0.0074 & 0.0262 $\pm$ 0.0040 & 0.0601 $\pm$ 0.0097 \\
DHT&0.6575 $\pm$ 0.0038 & 0.6680 $\pm$ 0.0088 & 0.0457 $\pm$ 0.0044 & 0.0696 $\pm$ 0.0089 \\
DSM&0.6718 $\pm$ 0.0033 & 0.6939 $\pm$ 0.0079 & 0.0315 $\pm$ 0.0047 & 0.0650 $\pm$ 0.0087 \\ \midrule
\rowcolor{Gray}DCM&0.6753 $\pm$ 0.0036 & 0.6939 $\pm$ 0.0079 & 0.0256 $\pm$ 0.0037 & 0.0561 $\pm$ 0.0085 \\ \midrule \bottomrule
\end{tabular}}
    
\caption{\small $C^{\text{td}}$ (higher means better discrimination) and ECE (lower means better calibration) of proposed approach versus baselines at the $75^{\text{th}}$ event quantile for SUPPORT.}
\label{fig:support_results_75th} 
\end{figure}

\begin{figure}[H]

\centering
\centerline{{\scriptsize SEER}}
    {\hspace{5em} \scriptsize Time Dependent Concordance Index \hfill Expected Calibration Error\hspace{7.5em}\phantom{} }\\

    \includegraphics[width=0.445\linewidth]{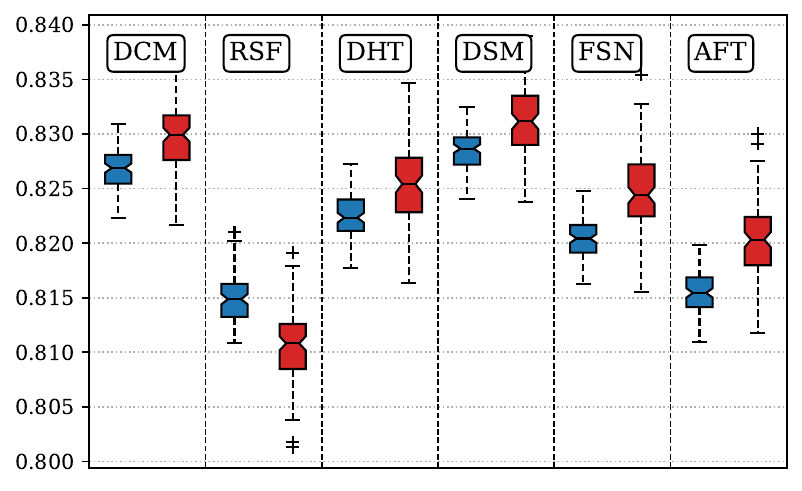}%
    \includegraphics[width=0.445\linewidth]{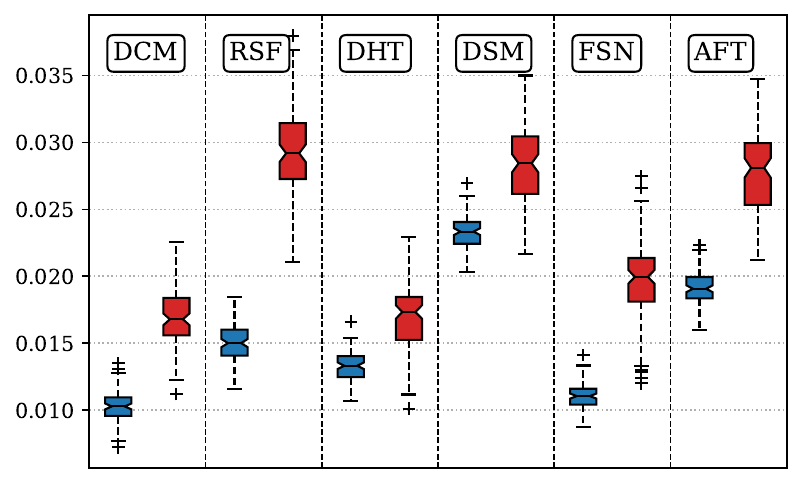}\\
    
    \resizebox{.875\textwidth}{!}{%
    \hspace{1em} \begin{tabular}{l|c|c|c|c}
    \toprule \midrule
     \multirow{2}{*}{ \textbf{Model}} & \multicolumn{2}{|c|}{\textbf{$C^{\text{td}}$}} & \multicolumn{2}{c|}{\textbf{ECE}}  \\ \cline{2-5} 
& {Population} & {Minority} & {Population} & {Minority}   \\ \hline
CPH&0.8082 $\pm$ 0.0020 & 0.8121 $\pm$ 0.0037 & 0.0718 $\pm$ 0.0015 & 0.0764 $\pm$ 0.0028 \\
AFT&0.8155 $\pm$ 0.0020 & 0.8204 $\pm$ 0.0035 & 0.0192 $\pm$ 0.0011 & 0.0278 $\pm$ 0.0029 \\
RSF&0.8153 $\pm$ 0.0021 & 0.8105 $\pm$ 0.0035 & 0.0147 $\pm$ 0.0013 & 0.0270 $\pm$ 0.0029  \\
FSN&0.8204 $\pm$ 0.0019 & 0.8248 $\pm$ 0.0036 & 0.0119 $\pm$ 0.0011 & 0.0196 $\pm$ 0.0029 \\
DHT&0.8224 $\pm$ 0.0020 & 0.8255 $\pm$ 0.0037 & 0.0133 $\pm$ 0.0012 & 0.0170 $\pm$ 0.0024 \\
DSM&0.8281 $\pm$ 0.0019 & 0.8243 $\pm$ 0.0036 & 0.0259 $\pm$ 0.0014 & 0.0311 $\pm$ 0.0031 \\ \midrule
\rowcolor{Gray}DCM&0.8270 $\pm$ 0.0019 & 0.8296 $\pm$ 0.0034 & 0.0103 $\pm$ 0.0011 & 0.0169 $\pm$ 0.0024 \\ \midrule \bottomrule
\end{tabular}}
    
\caption{\small $C^{\text{td}}$ (higher means better discrimination) and ECE (lower means better calibration) of proposed approach versus baselines at the $75^{\text{th}}$ event quantile for SEER.}
\label{fig:seer_results_75th} 
\end{figure}

In this section we describe the results of our various experiments with DCM and the competing baselines. We present the discriminative performance and calibration for DCM against the baselines on the three datasets for the entire population as well as the minority demographic on the $75^{\text{th}}$ quantile of event times in Figures \ref{fig:flchain_results_75th}, \ref{fig:support_results_75th} and \ref{fig:seer_results_75th} and the corresponding tables. (For tabulated results including AuROC and Brier Scores, refer to \ref{apx:results}.)\\

\noindent \textbf{FLCHAIN:} \model{} beat all the other baselines in terms of discriminative performance on the entire population as well as on the minority, `Female' subgroup. In terms of calibration \model{} was also consistently better than all the other baselines as evidenced from low ECE scores. Interestingly, both the FSN and the linear Cox model did poorly in terms of concordance and calibration while \model{} had good performance suggesting it is not sensistive to proportional hazards (PH). \\

\noindent \textbf{SUPPORT:} For the SUPPORT dataset, RSF had the best discriminative performance at a population level, and \model{} came a close and beat the other deep learning baselines. Interestingly we found that the proposed \model{} had the best discriminative performance on the minority demographic beating all other baselines including RSF. 

While RSF was strong in terms of discriminative performance, it was however poorly calibrated in comparison to other baselines. \model{} had the lowest ECE at each quantile amongst all baselines, at both the population level as well as on the minority demographic. The performance of FSN was close to \model{} in terms of calibration but did poorly in terms of discrimination, further lending evidence to the fact that DCM is not restricted by PH. \\

\noindent \textbf{SEER}: In terms of calibration  \model{} beat all the other baselines at all quantiles of interest for the entire population as well as for the minority group. \model{} also consistently had good discriminative power. (DSM did slightly better in terms of discrimination at a population level, but this was not significant). We found that DHT was a strong competitor, which is understandable since it is particularly well suited for discrete time datasets like SEER. Note that in the case of SEER we also report results stratified by the four largest minority demographics in the subset of the dataset we work with in Figure \ref{fig:seer_all_minorities}. \model{} has better discriminative performance across groups especially at longer horizons of event times. In terms of calibration, DCM comes close to or outperforms the semi-parametric approaches like FSN/DeepSurv.

In order to assess the influence of the protected attribute to determine the outcome we conduct additional studies of DCM on the SEER dataset involving removal of the protected attribute (unawareness). We find that unawareness results in overall poorer discriminative performance and calibration. Although unaware models were better calibrated at shorter time horizons, suggesting non-monotonic interaction of group attribute vis-a-vis calibration. These results are deferred to Appendix \ref{apx:unawareness}.

\subsection{Learnt Latent Groups}

For the SUPPORT dataset, we run DCM with the default set of hyperparameters with $k=3$ latent groups. We compare the subgroup level survival curves for the learnt subgroups using DCM by plotting the mean survival rates within each group estimated with DCM as well as a Kaplan Meier Estimator. 

\begin{figure}[!htbp]
    \centering
    \includegraphics[width=0.5\textwidth]{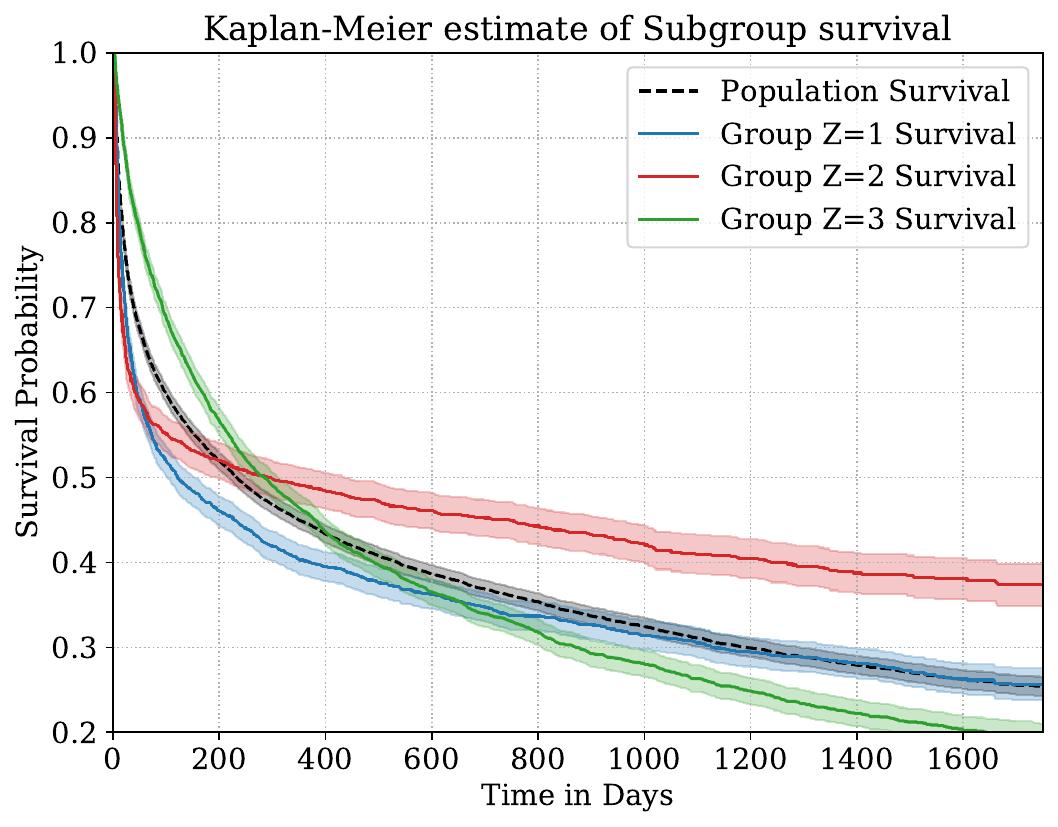}%
    \includegraphics[width=0.5\textwidth]{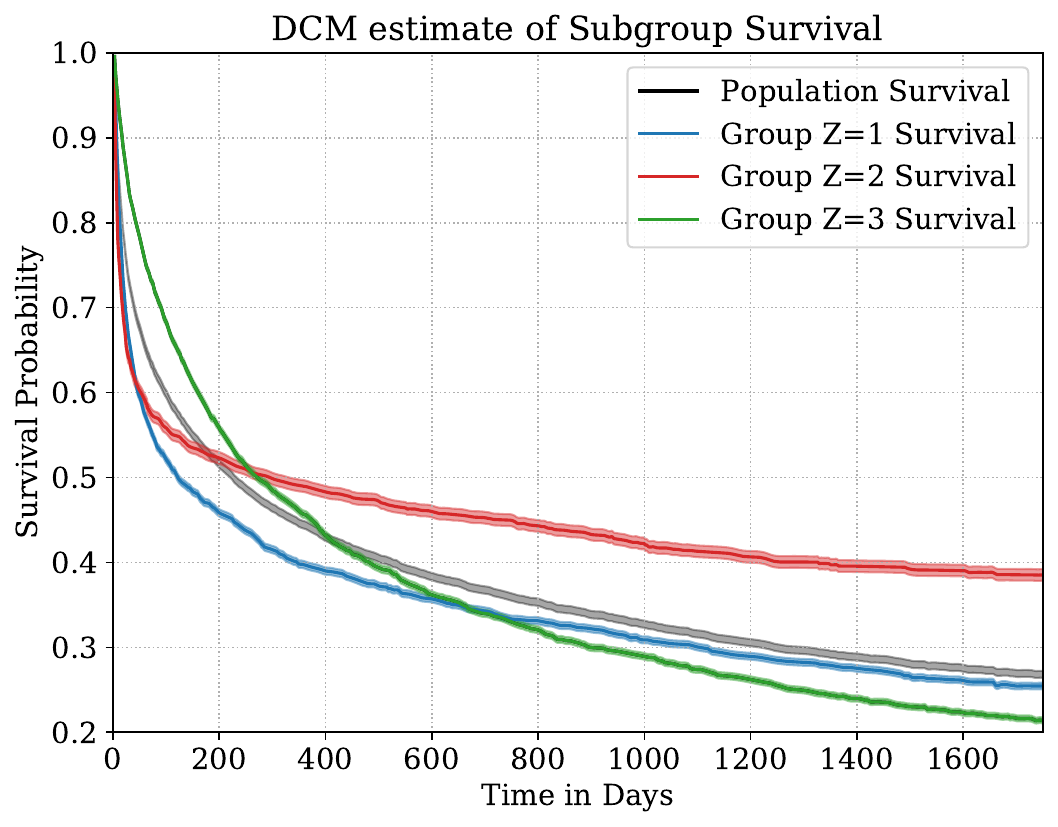}%

    \caption{Group specific baseline survival rates for the estimated subgroups using DCM with $k=3$ for a heldout fold of the SUPPORT dataset. The first plot is the group specific Kaplan-Meier plot and the second plot is the survival estimated with DCM. }
    \label{fig:dcmsubgroups}
\end{figure}

Figure \ref{fig:dcmsubgroups} present the estimate survival rates of the discovered subgroups using a Kaplan-Meier curve and DCM respectively. Consider the subgroup level survival curves for groups $Z=2$ and $Z=3$ that intersect. Intersecting survival curves indicate non Proportional Hazards which DCM is able to capture.

\section{Ethical considerations in survival prediction$^{*}$}
\label{apx:ethics}

Survival prediction and disease prognosis models can help on advanced care decision making and giving recommendations to the patients and their caregivers. Following the model card recommendations ~\citep{mitchell2019model}, we would like to discuss the intended uses of this technology, and that, using survival prediction techniques, healthcare providers can get more information on the treatment options, and in particular, preventive interventions and their potential outcomes given the predictive models and circumstantial characteristics. 

Our models are developed as a proof of concept and the existing datasets including the datasets used in these studies still lack some confounding characteristics that may be causally related to the final outcome~\citep{gaille2020ethical}. For example, studies show black women with breast cancer in the US have a higher mortality rate than white women~\citep{yedjou2019health}. However, there is an ongoing discussion on if the mortality rate is related to hereditary~\citep{yedjou2019health} or other factors such as late diagnosis due to the historical discrimination or a combination of all these factors~\citep{george2015diagnosis}. 
In addition, it is important to note that there should be extensive considerations in using survival analysis systems.

First, these systems should not be used to make decisions in a fully automated manner. There are many reasons for this including well documented biases embedded in historic data. These systems are also not intended to be used in cases of scarce resources, such as ventilators in Covid-19 cases, to rank patients by likelihood of survival\footnote{\url{https://www.healthaffairs.org/do/10.1377/hblog20200911.401376/}}~\citep{beil2019ethical}. Many people in the disability justice space have discussed how disabled patients are often discriminated against based on quality of life estimates~\cite{}. It is important to note that estimated survival rates, should not be the only means in decision making, and quality of life as a factor should be discussed with patients and their families.

In addition, it is very important to know where and how these predictions are used and who will have access to these analyses. For example, the outcome is not well-suited for usage in insurance policies premium~\citep{chiang1984life,czado2002application}, patient ranking and immigration status recommendations and not as an automatic means of decision making. 
In this work we address the issues of result disparity for underrepresented groups, however, we want to state that not all factors are presented in this study that may potentially contribute to the system predictions.
Finally, we stress that the causal association of protected attributes like ‘Race’ and ‘Gender’ with outcomes is largely problem dependent and an open research problem~\cite{}. Given that ‘Race’, ‘Gender’ are not clearly defined attributes even in medical contexts, we urge readers to exercise caution and best judgement when choosing to include these attributes to model outcomes.

\section{Conclusion}
We proposed `Deep Cox Mixtures' to model censored Time-to-Event data. Our approach involves estimating hazard ratios within latent clusters followed by non-parametric estimation of the baseline survival rates but is not limited by the strong assumptions of constant proportional hazards. We experiment with several real-world health datasets and demonstrate superiority of our approach both in terms of discriminative performance and calibration with an emphasis on improvements especially on the minority demographics.

In the future, we aim to apply Deep Cox Mixtures to real world health problems to phenotype and recover patients stratified by their relative risk profiles with the overall goal of actionable decision support for clinicians \citep{wang2017causal, ustun2019learning, chapfuwa2020survival}. Future extensions can also involve explicitly modelling the effect of an intervention or treatment at an individual \citep{chapfuwa2021enabling} or subgroup level \citep{nagpal2020interpretable} for retrospective analysis of studies with censored outcomes.


\bibliography{ref}

\newpage

\appendix

\noindent \centerline {\Large\textbf{ Supplementary Materials}}

\section{Additional details on \model{} implementation}
\label{apx:model}

\subsection{Non Applicability of the Partial Likelihood for the Proposed Model}
\label{apx:na}

In this section, we demonstrate that we cannot directly maximize the partial likelihood to learn our model. In the case of the Cox model, the hazard rate for an individual with covariates $\vx_i$ at time $t$, $\bm\lambda(t| \vx_i)$ is given as 
$$\bm\lambda(t| \vx_i) = \bm\lambda_0 (t) \exp(f(\beta, \vx_i)) .$$
Here, $\bm\lambda_0 (t)$ is the baseline hazard. Now the partial likelihood $\mathcal{PL}(\bm \theta)$ is defined as 
\begin{align}
\mathcal{PL}(\bm\theta) &= \prod_{i: \delta_i=1} \frac{\bm\lambda(t| \vx_i) }{\sum\limits_{j\in \mathcal{R}(t_i) } \bm\lambda(t| \vx_j) } = \prod_{i: \delta_i=1} \frac{ \bcancel{ \bm\lambda_0 (t) } \exp \big( f(\bm\theta; \vx_i) \big)}{\sum\limits_{j\in \mathcal{R}(t_i) } \bcancel{ \bm\lambda_0 (t) }  \exp \big( f(\bm\theta; \vx_j) \big)} \\  &= \prod_{i: \delta_i=1} \frac{ \exp \big( f(\bm\theta; \vx_i) \big)}{\sum\limits_{j\in \mathcal{R}(t_i) } \exp \big( f(\bm\theta; \vx_j) \big)}.
\end{align}
Under our model, the hazard rate for an individual with covariates $\vx_i$ at time $t$, $\bm\lambda(t| \vx_i)$ is given as 

$$\bm\lambda(\cdot| \vx_i) = \frac{\mathbb{P} (t|\vx_i) }{ \mS (t|\vx_i) } = \frac{ \sum\limits_k \mathbb{P} (t|\vx_i,Z=k) \mathbb{P} (Z=k|\vx_i) }{\sum\limits_k   \mS (t|\vx_i, Z=k)  \mathbb{P} (Z=k|\vx_i) } $$

Clearly, we do not have the proportional hazards form for DCM and so cannot directly optimize the Partial Likelihood independent of the baseline hazard rate.

\subsection{Spline Estimates}
\label{apx:model_splines}

 We want to extract the probabilities estimates $\mathbb{P}(T|Z,X)$ in order to compute the posterior $\mathbb{P}(Z|T, X) \propto \mathbb{P}(T|Z,X)$ for the uncensored observations. We only have access to the estimated survival function from the Breslow's estimate, $\widehat{\mS}(T>t|X=\vx_i) $. 
\begin{align*}
\mathbb{P}(T>t|X=\vx_i, Z=k)  &= 1 - \mathbb{P}(T\leq t|X=\vx_i, Z=k)\\
&= 1 - \textrm{cdf}(T\leq t|X=\vx_i, Z=k)\\
\text{Now, } \textrm{cdf}(T\leq t|X=\vx_i, Z=k) &= 1 - \mathbb{P}(T>t|X=\vx_i, Z=k)\\
\frac{ \partial }{\partial t}  \textrm{cdf}(T\leq t|X=\vx_i, Z=k) &= \frac{ \partial }{\partial t} \bigg( 1 - \mathbb{P}(T>t|X=\vx_i, Z=k) \bigg) \qquad \text{[taking derivative wrt. }t]\\
\implies \textrm{pdf}(T = t|X=\vx_i, Z=k) &=   - \frac{ \partial }{\partial t} \mathbb{P}(T>t|X=\vx_i, Z=k)\\
 &= - \frac{ \partial }{\partial t}  \bm{S}_k(t)^{\exp(f_k(\bm{\theta}; \vx_i ))}
\end{align*}
Here $\textrm{pdf}(\cdot)$ and $\textrm{cdf}(\cdot)$ are the probability density and the cumulative density functions respectively. Now replacing the baseline survival function $\bm{S}_k(.)$ with the interpolated spline estimate,  $\widetilde{\bm{S}}_k(.)$ we get the spline estimate of $\mathbb{P}(T=t|Z,X)$ as
\begin{align*}
 \widehat{\mathbb{P}}(T=t|Z,X) &=    - \frac{ \partial }{\partial t}  \widetilde{\bm{S}}_k(t)^{\exp(f_k(\bm{\theta}; \vx_i ))}\\
  &=  - \exp \big(f_k(\bm{\theta}; \vx_i ) \big)  {\widetilde{\mS}_k(t)}^{\exp \big(f_k(\bm{\theta}; \vx_i ) \big) -1  } \frac{ \partial }{\partial t}  \widetilde{\mS}_k(t) \\
 &=  - \exp \big(f_k(\bm{\theta}; \vx_i ) \big)   \frac{ \widehat{\mathbb{P}}(T>t|\vx_i, Z=k))}{\widetilde{\mS}_k(t)} \frac{ \partial }{\partial t}  \widetilde{\mS}_k(t) 
\end{align*}

Here, $\frac{ \partial }{\partial t}  \widetilde{\mS}_k(t) $ is the derivative of the baseline survival rate interpolated with a polynomial spline.

\section{Censoring adjusted evaluation metrics}
\label{apx:censoring}

\noindent \textbf{Area under ROC Curve} (AUC): The ROC curve is defined as a plot between the True Positive Rate/Sensitivity (TPR) and the False Positive Rate (FPR) for all thresholds at which a classifier can be deployed. Note that the FPR is equal to $1-$Specificity. We employ the technique proposed by \citesup{uno2007evaluating, hung2010optimal} to adjust the Sensitivity using IPCW estimates of the censoring distribution. The Specificity is computed on the uncensored instances.
$$\widehat{\text{Se}}(c, t)=\frac{\sum\limits_{i=1}^{n} \omega_i \cdot \mathbbm{1}\{ \pi_i(t) > c, T_i \leq t  \} \cdot   }{\sum\limits_{i=1}^{n} \omega_i \cdot \mathbbm{1}\{T_i < t  \} \cdot}; \quad \omega_i = \frac{\delta_i}{n\cdot \hat G(T_i)};\quad \widehat{\text{Sp}}(c, t)=\frac{\sum\limits_{i=1}^{n}  \mathbbm{1}\{ \pi_i(t) \leq c, T_i > t  \} \cdot   }{\sum\limits_{i=1}^{n} \mathbbm{1}\{T_i > t  \} \cdot}$$
$\widehat{\text{Se}}(c, t)$ and $\widehat{\text{Sp}}(c, t)$ refer to the estimated sensitivity and specificity at classification threshold $c$ and time horizon $t$ respectively. $\hat G(t)$ is a Kapaln-Meier estimator of the censoring distribution and $\pi_i(t)$ is the estimated survival probability, $\widehat{\mathbb{P}} (T>t|X=\vx_i)$ by the classifier. This curve is plotted for all thresholds $c \in [0,1]$ and the area under the curve is used to AUC. For a larger discussion around comparisons of various strategies to compute ROC curves in the presence of censoring refer to \citesup{kamarudin2017time}.\\

 \noindent\textbf{Time Dependent Concordance Index} ($C^\text{td}$): Concordance Index estimates ranking ability by exhaustively comparing relative risks across all pairs of individuals in the test set within a fixed horizon of time.
\begin{align*}
C^{td }(t) = \mathbb{P}\big( \pi_i(t) \leq \pi_j(t) | \delta_i=1, T_i<T_j, T_i \leq t \big) 
\end{align*}
Here, $\pi_i(t)$ is the estimated survival probability; $T$ represent the event times. In order to deal with censoring we employ the censoring adjusted estimator for $C^{\text{td}}$ that exploits IPCW estimates from a Kaplan-Meier estimate of the censoring distribution. The details are beyond the scope of this discussion and can be found in \citesup{uno2011c} and \citesup{gerds2013estimating}.\\

\noindent \textbf{Expected $\mathbf{\ell_1}$ Calibration Error }(ECE): The ECE measures the average absolute difference between the observed and expected (according to the risk score) event rates, conditional on the estimated risk score. At time $t$, let the predicted risk score be $R(t) = \widehat{\mathbb{P}}(T>t | X)$. Then, the ECE approximates
 \begin{align*}
\text{ECE}(t) = \mathbb{E} \big[ \big| \mathbb{P}(T > t | R(t)) - R(t)  \big| \big]
\end{align*}
by partitioning the risk scores $R$ into $q$ quantiles $\{[r_j, r_{j+1})\}_{j=1}^q$. and computing the Kaplan-Meier estimate of the event rate $\text{KM}_j(t) \approx P(T > t | R \in [r_j, r_{j+1}))$, and the average risk score $\overline{R}_j = \frac{q}{n} \sum_{i : R_i \in [r_j, r_{j+1})} R_i$ in each bin. Altogether, the estimated ECE is
\begin{equation*}
    \widehat{\text{ECE}}(t) = \frac{1}{q} \sum_{j = 1}^q |\text{KM}_j(t) - \overline{R}_j(t)|.
\end{equation*}
In practice, we fix the number of quantiles to be 20 for our experiments.\\

\noindent \textbf{Brier Score} (BS): The Brier Score involves computing the Mean Squared Error around the binary forecast of survival at a certain event quantile of interest. Brier Score is a proper scoring rule and can be decomposed into components that measure both discriminative performance and calibration.
\begin{align*}
\text{BS}(t) &= \mathbb{E}_{\mathcal{D}}\big[ \big(\mathbbm{1}\{ T_i > t \} - \widehat{\mathbb{P}}(T>t|X)\big)^2  \big]\\
\widehat{\text{BS}}_{\text{IPCW}}(t) &= \frac{1}{n}\sum_{i=1}^{n}\bigg[ \frac{ \pi_i(t)^2 \mathbbm{1}\{ T \leq t, \delta_i =1 \}}{\hat G_i(T_i)} +  \frac{ \big(1-\pi_i(t)\big)^2 \mathbbm{1}\{ T > t \}}{\hat G_i(t)} \bigg];\\ \text{where, } \pi_i(t) &= \widehat{\mathbb{P}}(T>t|X_i) 
\end{align*}
The adjusted Brier Score adjusted for Censoring using IPCW is given by  $\widehat{\text{BS}}_{\text{IPCW}}(t)$ as proposed in \citepsup{graf1999assessment, gerds2006consistent} Here, $\hat G(.)$ is the Kaplan Meier estimate of the Censoring Distribution. When the Censoring distribution is independent of the Event distribution, the above quantity is an unbiased estimate of the Brier Score.

\section{Hyper-Parameter tuning for the Baselines}
\label{apx:bl_hyper}
In this section we specify the hyper parameter choices along with a short description over which we perform grid search for the baselines. 
\begin{table}[!htbp]
    \centering
    \caption{DSM Hyper-parameter Grid }
\textbf{\texttt{
        \begin{tabular}{l|r}\toprule
        Hyper-parameter&Grid\\ \hline
         Outcome Distribution& \{ `Weibull' \}\\
         No. Clusters ($k$) &\{ `3', `4' \}\\
         No. of Hidden Layers &\{ `0', `1', `2' \}\\
         Hidden Layer Dim. &\{ `50', `100' \}\\
         Batch Size  &\{ `128', `256' \}\\
         Learning Rate  &\{ `1e-4', `1e-3' \}\\
         Activation&  \{ `SeLU' \}\\ \hline
         \bottomrule
    \end{tabular} }}
    \label{tab:hyper1}
\end{table}

\noindent \textbf{Deep Survival Machines (DSM)}: The choice of hyper parameters for DSM include the number of underlying survival distributions $(k)$ the choice of each outcome survival distribution, the number of hidden layers and neurons for the representation learning network and the activations. We also tune the learning rate and batch size. The choices of hyperparam values is given in Table \ref{tab:hyper1}. 
\begin{table}[!htbp]
    \centering
        \caption{DHT and FSN Hyper-parameter Grid }
\textbf{\texttt{
        \begin{tabular}{l|r}\toprule
        Hyper-parameter&Grid\\ \hline
         No. of Hidden Layers& \{ `1', `2' \}\\
         Hidden Layer Dim. &\{ `50', `100' \}\\
         Batch Size & \{ `128', `256' \}\\
         Learning Rate  &\{ `1e-4', `1e-3' \}\\
         Activation&  \{ `ReLU' \}\\ \hline
         \bottomrule
    \end{tabular} }}
    \label{tab:hyper2}
\end{table}

\noindent \textbf{Deep Hit (DHT)}: For Deep Hit, we tune the the Number of Hidden Layers, dimensionality of the hidden layers and the activation function. We also tune the learning rate and minibatch size. Note that Deep Hit requires grid discretization of the output event time space. For the SUPPORT and FLCHAIN datasets we discretize the output grid by dividing it into bins of ${\textrm{max}(T)}$ bins. Since, the SEER is a discrete event time dataset we divide the output grid for Deep Hit into $\nicefrac{\textrm{max}(T)}{10}$ bins.\\

\noindent \textbf{Faraggi-Simon Net (FSN)/DeepSurv}: Similar to Deep Hit, for FSN we tune the the Number of Hidden Layers, dimensionality of the hidden layers and the activation function. We also tune the learning rate and minibatch size. 

Both FSN and DHT were implemented using the \texttt{pycox} \citepsup{kvamme2019time} python package. Table \ref{tab:hyper2} describes the hyper-parameter choices for both DHT and FSN.

\begin{table}[!htbp]
    \caption{RSF Hyper-parameter Grid }
    \centering
\textbf{\texttt{
        \begin{tabular}{l|r}\toprule
        Hyper-parameter&Grid\\ \hline
         Max Depth& \{  `5' \}\\
         No. of Trees  &\{`50' \}\\
         mtry  &\{`sqrt' ,50 , 75, `all' \}\\ 
         min\_node\_split  &\{`150' , `200', `250' \}\\ 
         \bottomrule
    \end{tabular} }}
    \label{tab:hyper3}
\end{table}

\noindent \textbf{Random Survival Forest (RSF)}: For the RSF model we tune the number of trees and the maximum depth of each tree using the implementation as paart of the \texttt{pysurvival} Python package \citepsup{pysurvival_cite}. Table \ref{tab:hyper3} presents the chosen grid parameters.
\begin{table}[!htbp]
    \centering
    \caption{AFT and CPH Hyper-parameter Grid }
\textbf{\texttt{
        \begin{tabular}{l|r}\toprule
        Hyper-parameter&Grid\\ \hline
         $\ell 2$ Penalty& \{ `1e-3', `1e-2', `1e-1' \}\\
         \bottomrule
    \end{tabular} }}
    \label{tab:hyper4}
\end{table}

For \noindent \textbf{Cox Proportional Hazards (CPH)} and \noindent \textbf{Accelerated Failure Time (AFT)} the only hyperamaeter is the $\ell 2$ penalty on the parameters. The grid choice is presented in Table \ref{tab:hyper4}.

\section{Additional Results}
\subsection{Tabulated Results}
\label{apx:results}

In this section we present tabulated results for our experiments for the entire population and the minority demogrpahic on the three datasets.

\begin{table*}[!h]
    \centering
$C^{\text{td}}$($t$) ($\uparrow$)\\
\resizebox{0.685\textwidth}{!}{\begin{tabular}{l|c|c|c}
    \toprule \midrule
     \multirow{2}{*}{ \textbf{Model}} & \multicolumn{3}{|c|}{\textbf{Quantiles}}  \\ \cline{2-4} 
& {$t=25$th} & {$t=50$th} & {$t=75$th}   \\ \hline
CPH&0.6621 $\pm$ 0.0143 & 0.6696 $\pm$ 0.0110 & 0.6621 $\pm$ 0.0087 \\
AFT&0.7914 $\pm$ 0.0107 & 0.7938 $\pm$ 0.0080 & 0.7911 $\pm$ 0.0060 \\
RSF&0.7898 $\pm$ 0.0102 & 0.7908 $\pm$ 0.0078 & 0.7880 $\pm$ 0.0059 \\
FSN&0.6353 $\pm$ 0.0146 & 0.6519 $\pm$ 0.0104 & 0.6608 $\pm$ 0.0081 \\
DSM&0.8008 $\pm$ 0.0100 & 0.7988 $\pm$ 0.0078 & 0.7937 $\pm$ 0.0061 \\
DHT&0.7669 $\pm$ 0.0104 & 0.7666 $\pm$ 0.0078 & 0.7636 $\pm$ 0.0059 \\ \midrule
\rowcolor{Gray}DCM&0.7991 $\pm$ 0.0103 & 0.7988 $\pm$ 0.0077 & 0.7943 $\pm$ 0.0060 \\ 
\midrule \bottomrule
\end{tabular}}

\vspace{1em}
    \textsc{AUC($t$)} ($\uparrow$)\\
\resizebox{0.685\textwidth}{!}{\begin{tabular}{l|c|c|c}
    \toprule \midrule
     \multirow{2}{*}{ \textbf{Model}} & \multicolumn{3}{|c|}{\textbf{Quantiles}}  \\ \cline{2-4} 
& {$t=25$th} & {$t=50$th} & {$t=75$th}   \\ \hline
CPH&0.6680 $\pm$ 0.0149 & 0.6827 $\pm$ 0.0120 & 0.6821 $\pm$ 0.0094 \\
AFT&0.8032 $\pm$ 0.0110 & 0.8170 $\pm$ 0.0085 & 0.8257 $\pm$ 0.0063 \\
RSF&0.8015 $\pm$ 0.0105 & 0.8142 $\pm$ 0.0083 & 0.8235 $\pm$ 0.0064 \\
FSN&0.6416 $\pm$ 0.0150 & 0.6673 $\pm$ 0.0109 & 0.6904 $\pm$ 0.0090 \\
DSM&0.8124 $\pm$ 0.0102 & 0.8218 $\pm$ 0.0083 & 0.8283 $\pm$ 0.0066 \\
DHT&0.7771 $\pm$ 0.0106 & 0.7878 $\pm$ 0.0082 & 0.7936 $\pm$ 0.0064 \\ \midrule
\rowcolor{Gray}DCM&0.8107 $\pm$ 0.0106 & 0.8219 $\pm$ 0.0082 & 0.8291 $\pm$ 0.0065 \\
\midrule \bottomrule
\end{tabular}}

\vspace{1em}
  \textsc{ECE($t$)} ($\downarrow$)\\
\resizebox{0.685\textwidth}{!}{\begin{tabular}{l|c|c|c}
    \toprule \midrule
     \multirow{2}{*}{ \textbf{Model}} & \multicolumn{3}{|c|}{\textbf{Quantiles}}  \\ \cline{2-4} 
& {$t=25$th} & {$t=50$th} & {$t=75$th}   \\ \hline
CPH&0.0386 $\pm$ 0.0031 & 0.0699 $\pm$ 0.0042 & 0.0992 $\pm$ 0.0044 \\
AFT&0.0141 $\pm$ 0.0024 & 0.0216 $\pm$ 0.0034 & 0.0212 $\pm$ 0.0034 \\
RSF&0.0155 $\pm$ 0.0022 & 0.0198 $\pm$ 0.0027 & 0.0215 $\pm$ 0.0037 \\
FSN&0.0214 $\pm$ 0.0027 & 0.0334 $\pm$ 0.0035 & 0.0381 $\pm$ 0.0046 \\
DSM&0.0144 $\pm$ 0.0025 & 0.0214 $\pm$ 0.0030 & 0.0223 $\pm$ 0.0029 \\
DHT&0.0283 $\pm$ 0.0029 & 0.0410 $\pm$ 0.0036 & 0.0505 $\pm$ 0.0041 \\ \midrule
\rowcolor{Gray}DCM&0.0122 $\pm$ 0.0024 & 0.0169 $\pm$ 0.0033 & 0.0200 $\pm$ 0.0034 \\
\midrule \bottomrule 
\end{tabular}}

\vspace{1em}
BS($t$)($\downarrow$)\\
\resizebox{0.685\textwidth}{!}{\begin{tabular}{l|c|c|c}
    \toprule \midrule
     \multirow{2}{*}{ \textbf{Model}} & \multicolumn{3}{|c|}{\textbf{Quantiles}}  \\ \cline{2-4} 
& {$t=25$th} & {$t=50$th} & {$t=75$th}   \\ \hline
CPH&0.0671 $\pm$ 0.0027 & 0.1211 $\pm$ 0.0035 & 0.1665 $\pm$ 0.0037 \\
AFT&0.0584 $\pm$ 0.0023 & 0.0991 $\pm$ 0.0028 & 0.1244 $\pm$ 0.0025 \\
RSF&0.0603 $\pm$ 0.0023 & 0.1004 $\pm$ 0.0027 & 0.1250 $\pm$ 0.0026 \\
FSN&0.0672 $\pm$ 0.0026 & 0.1199 $\pm$ 0.0029 & 0.1589 $\pm$ 0.0027 \\
DSM&0.0578 $\pm$ 0.0022 & 0.0975 $\pm$ 0.0028 & 0.1224 $\pm$ 0.0026 \\
DHT&0.0631 $\pm$ 0.0022 & 0.1086 $\pm$ 0.0026 & 0.1399 $\pm$ 0.0024 \\ \midrule
\rowcolor{Gray}DCM&0.0582 $\pm$ 0.0023 & 0.0979 $\pm$ 0.0028 & 0.1228 $\pm$ 0.0026 \\  
\midrule \bottomrule
\end{tabular}}

    \caption{\small Results for various performance metrics on FLCHAIN (entire population) along with bootstrapped std errors.}
    \label{tab:res_flchain}
\end{table*}

\begin{table*}[!ht]
    \centering
$C^{\text{td}}$($t$) ($\uparrow$)\\
\resizebox{0.685\textwidth}{!}{\begin{tabular}{l|c|c|c}
    \toprule \midrule
     \multirow{2}{*}{ \textbf{Model}} & \multicolumn{3}{|c|}{\textbf{Quantiles}}  \\ \cline{2-4} 
& {$t=25$th} & {$t=50$th} & {$t=75$th}   \\ \hline
CPH&0.6444 $\pm$ 0.0193 & 0.6692 $\pm$ 0.0160 & 0.6737 $\pm$ 0.0124 \\
AFT&0.7822 $\pm$ 0.0158 & 0.7838 $\pm$ 0.0112 & 0.7875 $\pm$ 0.0087 \\
RSF&0.7796 $\pm$ 0.0147 & 0.7799 $\pm$ 0.0113 & 0.7830 $\pm$ 0.0089 \\
FSN&0.5746 $\pm$ 0.0211 & 0.6014 $\pm$ 0.0156 & 0.6212 $\pm$ 0.0131 \\
DSM&0.7849 $\pm$ 0.0153 & 0.7886 $\pm$ 0.0113 & 0.7909 $\pm$ 0.0087 \\
DHT&0.7607 $\pm$ 0.0153 & 0.7610 $\pm$ 0.0116 & 0.7631 $\pm$ 0.0092 \\ \midrule
\rowcolor{Gray}DCM&0.7873 $\pm$ 0.0164 & 0.7893 $\pm$ 0.0116 & 0.7911 $\pm$ 0.0091 \\
\midrule \bottomrule
\end{tabular}}

\vspace{1em}
\textsc{AUC($t$)} ($\uparrow$)\\
\resizebox{0.685\textwidth}{!}{\begin{tabular}{l|c|c|c}
    \toprule \midrule
     \multirow{2}{*}{ \textbf{Model}} & \multicolumn{3}{|c|}{\textbf{Quantiles}}  \\ \cline{2-4} 
& {$t=25$th} & {$t=50$th} & {$t=75$th}   \\ \hline
CPH&0.6492 $\pm$ 0.0202 & 0.6842 $\pm$ 0.0175 & 0.6983 $\pm$ 0.0136 \\
AFT&0.7944 $\pm$ 0.0163 & 0.8069 $\pm$ 0.0122 & 0.8230 $\pm$ 0.0095 \\
RSF&0.7918 $\pm$ 0.0152 & 0.8028 $\pm$ 0.0124 & 0.8189 $\pm$ 0.0099 \\
FSN&0.5774 $\pm$ 0.0219 & 0.6115 $\pm$ 0.0164 & 0.6477 $\pm$ 0.0148 \\
DSM&0.7966 $\pm$ 0.0158 & 0.8118 $\pm$ 0.0123 & 0.8259 $\pm$ 0.0095 \\
DHT&0.7710 $\pm$ 0.0157 & 0.7822 $\pm$ 0.0127 & 0.7938 $\pm$ 0.0104 \\ \midrule
\rowcolor{Gray}DCM&0.7991 $\pm$ 0.0169 & 0.8122 $\pm$ 0.0126 & 0.8265 $\pm$ 0.0100 \\\midrule \bottomrule
\end{tabular}}

\vspace{1em}
  \textsc{ECE($t$)} ($\downarrow$)\\
\resizebox{0.685\textwidth}{!}{\begin{tabular}{l|c|c|c}
    \toprule \midrule
     \multirow{2}{*}{ \textbf{Model}} & \multicolumn{3}{|c|}{\textbf{Quantiles}}  \\ \cline{2-4} 
& {$t=25$th} & {$t=50$th} & {$t=75$th}   \\ \hline
CPH&0.0378 $\pm$ 0.0044 & 0.0642 $\pm$ 0.0056 & 0.0878 $\pm$ 0.0071 \\
AFT&0.0221 $\pm$ 0.0035 & 0.0289 $\pm$ 0.0045 & 0.0329 $\pm$ 0.0046 \\
RSF&0.0220 $\pm$ 0.0036 & 0.0330 $\pm$ 0.0046 & 0.0368 $\pm$ 0.0053 \\
FSN&0.0325 $\pm$ 0.0043 & 0.0416 $\pm$ 0.0059 & 0.0545 $\pm$ 0.0068 \\
DSM&0.0243 $\pm$ 0.0038 & 0.0323 $\pm$ 0.0048 & 0.0347 $\pm$ 0.0056 \\
DHT&0.0328 $\pm$ 0.0037 & 0.0411 $\pm$ 0.0051 & 0.0525 $\pm$ 0.0056 \\ \midrule
\rowcolor{Gray}DCM&0.0209 $\pm$ 0.0035 & 0.0298 $\pm$ 0.0054 & 0.0294 $\pm$ 0.0049 \\
\midrule \bottomrule
\end{tabular}}

\vspace{1em}

BS($t$)($\downarrow$)\\
\resizebox{0.685\textwidth}{!}{\begin{tabular}{l|c|c|c}
    \toprule \midrule
     \multirow{2}{*}{ \textbf{Model}} & \multicolumn{3}{|c|}{\textbf{Quantiles}}  \\ \cline{2-4} 
& {$t=25$th} & {$t=50$th} & {$t=75$th}   \\ \hline
CPH&0.0693 $\pm$ 0.0043 & 0.1223 $\pm$ 0.0053 & 0.1626 $\pm$ 0.0057 \\
AFT&0.0613 $\pm$ 0.0035 & 0.1031 $\pm$ 0.0040 & 0.1262 $\pm$ 0.0037 \\
RSF&0.0624 $\pm$ 0.0036 & 0.1041 $\pm$ 0.0041 & 0.1273 $\pm$ 0.0038 \\
FSN&0.0715 $\pm$ 0.0043 & 0.1278 $\pm$ 0.0051 & 0.1673 $\pm$ 0.0050 \\
DSM&0.0609 $\pm$ 0.0035 & 0.1015 $\pm$ 0.0041 & 0.1244 $\pm$ 0.0037 \\
DHT&0.0648 $\pm$ 0.0035 & 0.1107 $\pm$ 0.0038 & 0.1394 $\pm$ 0.0039 \\ \midrule
\rowcolor{Gray}DCM&0.0607 $\pm$ 0.0035 & 0.1021 $\pm$ 0.0042 & 0.1249 $\pm$ 0.0039\\ \midrule \bottomrule
\end{tabular}}

    \caption{\small Results for various performance metrics on FLCHAIN (minority) along with bootstrapped std errors.}
    \label{tab:res_flchain_min}
\end{table*}

Tables \ref{tab:res_flchain} and \ref{tab:res_flchain_min} present the $C^{\text{td}}$, AUC, ECE and Brier Score for the Entire Population and Minority Demographic on the FLCHAIN dataset, respectively.

\begin{table*}[!h]
    \centering
    
$C^{\text{td}}$($t$) ($\uparrow$)\\
\resizebox{0.685\textwidth}{!}{\begin{tabular}{l|c|c|c}
    \toprule \midrule
     \multirow{2}{*}{ \textbf{Model}} & \multicolumn{3}{|c|}{\textbf{Quantiles}}  \\ \cline{2-4} 
& {$t=25$th} & {$t=50$th} & {$t=75$th}   \\ \hline
CPH&0.6899 $\pm$ 0.0057 & 0.6713 $\pm$ 0.0040 & 0.6686 $\pm$ 0.0034 \\
AFT&0.6826 $\pm$ 0.0057 & 0.6662 $\pm$ 0.0040 & 0.6657 $\pm$ 0.0034 \\
RSF&0.7513 $\pm$ 0.0063 & 0.7104 $\pm$ 0.0045 & 0.6751 $\pm$ 0.0040 \\
FSN&0.6988 $\pm$ 0.0059 & 0.6779 $\pm$ 0.0044 & 0.6736 $\pm$ 0.0037 \\
DSM&0.7459 $\pm$ 0.0059 & 0.7042 $\pm$ 0.0038 & 0.6718 $\pm$ 0.0033 \\
DHT&0.7302 $\pm$ 0.0067 & 0.6871 $\pm$ 0.0043 & 0.6575 $\pm$ 0.0038 \\ \midrule
\rowcolor{Gray}DCM&0.7425 $\pm$ 0.0059 & 0.7057 $\pm$ 0.0042 & 0.6753 $\pm$ 0.0036  \\ 
\midrule \bottomrule
\end{tabular}}

\vspace{1em}
    \textsc{AUC($t$)} ($\uparrow$)\\
\resizebox{0.685\textwidth}{!}{\begin{tabular}{l|c|c|c}
    \toprule \midrule
     \multirow{2}{*}{ \textbf{Model}} & \multicolumn{3}{|c|}{\textbf{Quantiles}}  \\ \cline{2-4} 
& {$t=25$th} & {$t=50$th} & {$t=75$th}   \\ \hline
CPH&0.7011 $\pm$ 0.0061 & 0.6990 $\pm$ 0.0049 & 0.7214 $\pm$ 0.0049 \\
AFT&0.6936 $\pm$ 0.0061 & 0.6943 $\pm$ 0.0049 & 0.7209 $\pm$ 0.0049 \\
RSF&0.7663 $\pm$ 0.0066 & 0.7379 $\pm$ 0.0054 & 0.7273 $\pm$ 0.0054 \\
FSN&0.7091 $\pm$ 0.0062 & 0.7050 $\pm$ 0.0052 & 0.7249 $\pm$ 0.0050 \\
DSM&0.7606 $\pm$ 0.0063 & 0.7337 $\pm$ 0.0047 & 0.7236 $\pm$ 0.0050 \\
DHT&0.7421 $\pm$ 0.0070 & 0.7123 $\pm$ 0.0052 & 0.7042 $\pm$ 0.0052 \\ \midrule
\rowcolor{Gray}DCM&0.7576 $\pm$ 0.0065 & 0.7347 $\pm$ 0.0049 & 0.7256 $\pm$ 0.0054 \\
\midrule \bottomrule
\end{tabular}}

\vspace{1em}
  \textsc{ECE($t$)} ($\downarrow$)\\
\resizebox{0.685\textwidth}{!}{\begin{tabular}{l|c|c|c}
    \toprule \midrule
     \multirow{2}{*}{ \textbf{Model}} & \multicolumn{3}{|c|}{\textbf{Quantiles}}  \\ \cline{2-4} 
& {$t=25$th} & {$t=50$th} & {$t=75$th}   \\ \hline
CPH&0.0201 $\pm$ 0.0029 & 0.0265 $\pm$ 0.0038 & 0.0310 $\pm$ 0.0041 \\
AFT&0.0281 $\pm$ 0.0031 & 0.0617 $\pm$ 0.0048 & 0.0402 $\pm$ 0.0046 \\
RSF&0.0241 $\pm$ 0.0032 & 0.0368 $\pm$ 0.0044 & 0.0348 $\pm$ 0.0041 \\
FSN&0.0220 $\pm$ 0.0029 & 0.0267 $\pm$ 0.0036 & 0.0262 $\pm$ 0.0040 \\
DSM&0.0341 $\pm$ 0.0033 & 0.0621 $\pm$ 0.0043 & 0.0315 $\pm$ 0.0047 \\
DHT&0.0220 $\pm$ 0.0026 & 0.0351 $\pm$ 0.0037 & 0.0457 $\pm$ 0.0044 \\ \midrule
\rowcolor{Gray}DCM&0.0179 $\pm$ 0.0030 & 0.0268 $\pm$ 0.0038 & 0.0256 $\pm$ 0.0037 \\
 \midrule \bottomrule
\end{tabular}}

\vspace{1em}
BS($t$) ($\downarrow$)\\
\resizebox{0.685\textwidth}{!}{\begin{tabular}{l|c|c|c}
    \toprule \midrule
     \multirow{2}{*}{ \textbf{Model}} & \multicolumn{3}{|c|}{\textbf{Quantiles}}  \\ \cline{2-4} 
& {$t=25$th} & {$t=50$th} & {$t=75$th}   \\ \hline
CPH&0.1334 $\pm$ 0.0023 & 0.1995 $\pm$ 0.0019 & 0.2136 $\pm$ 0.0016 \\
AFT&0.1354 $\pm$ 0.0025 & 0.2051 $\pm$ 0.0023 & 0.2147 $\pm$ 0.0016 \\
RSF&0.1240 $\pm$ 0.0023 & 0.1899 $\pm$ 0.0018 & 0.2109 $\pm$ 0.0017 \\
FSN&0.1315 $\pm$ 0.0023 & 0.1981 $\pm$ 0.0020 & 0.2122 $\pm$ 0.0018 \\
DSM&0.1271 $\pm$ 0.0024 & 0.1955 $\pm$ 0.0022 & 0.2130 $\pm$ 0.0017 \\
DHT&0.1271 $\pm$ 0.0024 & 0.1971 $\pm$ 0.0016 & 0.2206 $\pm$ 0.0014 \\ \midrule
\rowcolor{Gray}DCM&0.1258 $\pm$ 0.0024 & 0.1905 $\pm$ 0.0020 & 0.2118 $\pm$ 0.0019\\
\midrule \bottomrule
\end{tabular}}

    \caption{\small Results for various performance metrics on SUPPORT (entire population) along with bootstrapped std. errors.}
    \label{tab:res_support}
\end{table*}

\begin{table*}[!t]
    \centering
    \vspace{1em}
$C^{\text{td}}$($t$) ($\uparrow$)\\
\resizebox{0.685\textwidth}{!}{\begin{tabular}{l|c|c|c}
    \toprule \midrule
     \multirow{2}{*}{ \textbf{Model}} & \multicolumn{3}{|c|}{\textbf{Quantiles}}  \\ \cline{2-4} 
& {$t=25$th} & {$t=50$th} & {$t=75$th}   \\ \hline
CPH&0.7161 $\pm$ 0.0126 & 0.6982 $\pm$ 0.0089 & 0.6905 $\pm$ 0.0078 \\
AFT&0.7101 $\pm$ 0.0126 & 0.6941 $\pm$ 0.0089 & 0.6883 $\pm$ 0.0078 \\
RSF&0.7503 $\pm$ 0.0120 & 0.7198 $\pm$ 0.0084 & 0.6974 $\pm$ 0.0084 \\
FSN&0.7203 $\pm$ 0.0129 & 0.7025 $\pm$ 0.0090 & 0.6961 $\pm$ 0.0074 \\
DSM&0.7548 $\pm$ 0.0132 & 0.7220 $\pm$ 0.0093 & 0.6939 $\pm$ 0.0079 \\
DHT&0.7321 $\pm$ 0.0145 & 0.6943 $\pm$ 0.0099 & 0.6680 $\pm$ 0.0088 \\ \midrule
\rowcolor{Gray}DCM&0.7570 $\pm$ 0.0130 & 0.7234 $\pm$ 0.0089 & 0.6939 $\pm$ 0.0079 \\
\midrule \bottomrule
\end{tabular}}
    
    \vspace{1em}

    \textsc{AUC($t$)} ($\uparrow$)\\
\resizebox{0.685\textwidth}{!}{\begin{tabular}{l|c|c|c}
    \toprule \midrule
     \multirow{2}{*}{ \textbf{Model}} & \multicolumn{3}{|c|}{\textbf{Quantiles}}  \\ \cline{2-4} 
& {$t=25$th} & {$t=50$th} & {$t=75$th}   \\ \hline
CPH&0.7261 $\pm$ 0.0127 & 0.7348 $\pm$ 0.0109 & 0.7446 $\pm$ 0.0107 \\
AFT&0.7199 $\pm$ 0.0127 & 0.7311 $\pm$ 0.0109 & 0.7446 $\pm$ 0.0108 \\
RSF&0.7667 $\pm$ 0.0121 & 0.7536 $\pm$ 0.0106 & 0.7522 $\pm$ 0.0122 \\
FSN&0.7283 $\pm$ 0.0128 & 0.7375 $\pm$ 0.0110 & 0.7518 $\pm$ 0.0101 \\
DSM&0.7690 $\pm$ 0.0130 & 0.7594 $\pm$ 0.0113 & 0.7478 $\pm$ 0.0109 \\
DHT&0.7400 $\pm$ 0.0143 & 0.7265 $\pm$ 0.0123 & 0.7129 $\pm$ 0.0120 \\ \midrule
\rowcolor{Gray}DCM&0.7701 $\pm$ 0.0129 & 0.7588 $\pm$ 0.0109 & 0.7424 $\pm$ 0.0113 \\ 
\midrule \bottomrule
\end{tabular}}

\vspace{1em}

  \textsc{ECE($t$)} ($\downarrow$)\\
\resizebox{0.685\textwidth}{!}{\begin{tabular}{l|c|c|c}
    \toprule \midrule
     \multirow{2}{*}{ \textbf{Model}} & \multicolumn{3}{|c|}{\textbf{Quantiles}}  \\ \cline{2-4} 
& {$t=25$th} & {$t=50$th} & {$t=75$th}   \\ \hline
CPH&0.0473 $\pm$ 0.0071 & 0.0610 $\pm$ 0.0084 & 0.0685 $\pm$ 0.0079 \\
AFT&0.0530 $\pm$ 0.0075 & 0.0891 $\pm$ 0.0091 & 0.0741 $\pm$ 0.0085 \\
RSF&0.0401 $\pm$ 0.0064 & 0.0608 $\pm$ 0.0077 & 0.0603 $\pm$ 0.0080 \\
FSN&0.0418 $\pm$ 0.0067 & 0.0579 $\pm$ 0.0090 & 0.0601 $\pm$ 0.0097 \\
DSM&0.0506 $\pm$ 0.0070 & 0.0818 $\pm$ 0.0094 & 0.0650 $\pm$ 0.0087 \\
DHT&0.0483 $\pm$ 0.0070 & 0.0635 $\pm$ 0.0087 & 0.0696 $\pm$ 0.0089 \\ \midrule
\rowcolor{Gray}DCM&0.0397 $\pm$ 0.0059 & 0.0550 $\pm$ 0.0080 & 0.0561 $\pm$ 0.0085 \\
 \midrule \bottomrule
\end{tabular}}

\vspace{1em}
BS($t$) ($\downarrow$)\\
\resizebox{0.685\textwidth}{!}{\begin{tabular}{l|c|c|c}
    \toprule \midrule
     \multirow{2}{*}{ \textbf{Model}} & \multicolumn{3}{|c|}{\textbf{Quantiles}}  \\ \cline{2-4} 
& {$t=25$th} & {$t=50$th} & {$t=75$th}   \\ \hline
CPH&0.1340 $\pm$ 0.0050 & 0.1943 $\pm$ 0.0042 & 0.2069 $\pm$ 0.0037 \\
AFT&0.1363 $\pm$ 0.0054 & 0.2026 $\pm$ 0.0049 & 0.2090 $\pm$ 0.0039 \\
RSF&0.1263 $\pm$ 0.0048 & 0.1870 $\pm$ 0.0039 & 0.2031 $\pm$ 0.0039 \\
FSN&0.1319 $\pm$ 0.0051 & 0.1934 $\pm$ 0.0044 & 0.2037 $\pm$ 0.0040 \\
DSM&0.1275 $\pm$ 0.0050 & 0.1919 $\pm$ 0.0047 & 0.2056 $\pm$ 0.0040 \\
DHT&0.1298 $\pm$ 0.0049 & 0.1963 $\pm$ 0.0039 & 0.2186 $\pm$ 0.0036 \\ \midrule
\rowcolor{Gray}DCM&0.1261 $\pm$ 0.0048 & 0.1868 $\pm$ 0.0044 & 0.2073 $\pm$ 0.0044 \\ 
\midrule \bottomrule
\end{tabular}}

    \caption{\small Results for various performance metrics on SUPPORT (minority) along with bootstrapped std. errors.}
    \label{tab:res_support_min}
\end{table*}

Tables \ref{tab:res_support} and \ref{tab:res_support_min} present the $C^{\text{td}}$, AUC, ECE and Brier Score for the Entire Population and Minority Demographic on the SUPPORT dataset, respectively.


\begin{table*}[!t]
    \centering
    
    \vspace{1em}
$C^{\text{td}}$($t$) ($\uparrow$)\\
\resizebox{0.685\textwidth}{!}{\begin{tabular}{l|c|c|c}
    \toprule \midrule
     \multirow{2}{*}{ \textbf{Model}} & \multicolumn{3}{|c|}{\textbf{Quantiles}}  \\ \cline{2-4} 
& {$t=25$th} & {$t=50$th} & {$t=75$th}   \\ \hline
CPH&0.8766 $\pm$ 0.0027 & 0.8354 $\pm$ 0.0024 & 0.8082 $\pm$ 0.0020 \\
AFT&0.8823 $\pm$ 0.0026 & 0.8416 $\pm$ 0.0024 & 0.8155 $\pm$ 0.0020 \\
RSF&0.8838 $\pm$ 0.0025 & 0.8421 $\pm$ 0.0025 & 0.8153 $\pm$ 0.0021 \\
FSN&0.8850 $\pm$ 0.0025 & 0.8447 $\pm$ 0.0023 & 0.8204 $\pm$ 0.0019 \\
DHT&0.8915 $\pm$ 0.0024 & 0.8517 $\pm$ 0.0024 & 0.8224 $\pm$ 0.0020 \\
DSM&0.8949 $\pm$ 0.0022 & 0.8559 $\pm$ 0.0022 & 0.8281 $\pm$ 0.0019 \\ \midrule
\rowcolor{Gray}DCM&0.8933 $\pm$ 0.0024 & 0.8550 $\pm$ 0.0022 & 0.8270 $\pm$ 0.0019 \\\bottomrule
\end{tabular}}
    
    \vspace{1em}

    \textsc{AUC($t$)} ($\uparrow$)\\
\resizebox{0.685\textwidth}{!}{\begin{tabular}{l|c|c|c}
    \toprule \midrule
     \multirow{2}{*}{ \textbf{Model}} & \multicolumn{3}{|c|}{\textbf{Quantiles}}  \\ \cline{2-4} 
& {$t=25$th} & {$t=50$th} & {$t=75$th}   \\ \hline
CPH&0.8828 $\pm$ 0.0028 & 0.8526 $\pm$ 0.0025 & 0.8337 $\pm$ 0.0022 \\
AFT&0.8893 $\pm$ 0.0026 & 0.8596 $\pm$ 0.0025 & 0.8424 $\pm$ 0.0021 \\
RSF&0.8899 $\pm$ 0.0026 & 0.8594 $\pm$ 0.0026 & 0.8416 $\pm$ 0.0023 \\
FSN&0.8921 $\pm$ 0.0026 & 0.8632 $\pm$ 0.0024 & 0.8477 $\pm$ 0.0021 \\
DHT&0.8983 $\pm$ 0.0025 & 0.8701 $\pm$ 0.0025 & 0.8495 $\pm$ 0.0022 \\
DSM&0.9022 $\pm$ 0.0023 & 0.8748 $\pm$ 0.0023 & 0.8566 $\pm$ 0.0020 \\ \midrule
\rowcolor{Gray}DCM&0.9002 $\pm$ 0.0025 & 0.87350 $\pm$ 0.0023 & 0.8552 $\pm$ 0.0020\\ \midrule \bottomrule
\end{tabular}}

\vspace{1em}

  \textsc{ECE($t$)} ($\downarrow$)\\
\resizebox{0.685\textwidth}{!}{\begin{tabular}{l|c|c|c}
    \toprule \midrule
     \multirow{2}{*}{ \textbf{Model}} & \multicolumn{3}{|c|}{\textbf{Quantiles}}  \\ \cline{2-4} 
& {$t=25$th} & {$t=50$th} & {$t=75$th}   \\ \hline
CPH&0.0356 $\pm$ 0.0008 & 0.0577 $\pm$ 0.0012 & 0.0718 $\pm$ 0.0015 \\
AFT&0.0168 $\pm$ 0.0008 & 0.0187 $\pm$ 0.0011 & 0.0192 $\pm$ 0.0011 \\
RSF&0.0052 $\pm$ 0.0007 & 0.0092 $\pm$ 0.0010 & 0.0147 $\pm$ 0.0013 \\
FSN&0.0124 $\pm$ 0.0008 & 0.0140 $\pm$ 0.0011 & 0.0111 $\pm$ 0.0011 \\
DHT&0.0076 $\pm$ 0.0008 & 0.0115 $\pm$ 0.0011 & 0.0133 $\pm$ 0.0012 \\
DSM&0.0067 $\pm$ 0.0007 & 0.0211 $\pm$ 0.0012 & 0.0259 $\pm$ 0.0014 \\ \midrule
\rowcolor{Gray}DCM&0.0055 $\pm$ 0.0008 & 0.0087 $\pm$ 0.0010 & 0.0103 $\pm$ 0.0011 \\ \bottomrule
\end{tabular}}

\vspace{1em}
BS($t$) ($\downarrow$)\\
\resizebox{0.685\textwidth}{!}{\begin{tabular}{l|c|c|c}
    \toprule \midrule
     \multirow{2}{*}{ \textbf{Model}} & \multicolumn{3}{|c|}{\textbf{Quantiles}}  \\ \cline{2-4} 
& {$t=25$th} & {$t=50$th} & {$t=75$th}   \\ \hline
CPH&0.0501 $\pm$ 0.0007 & 0.0887 $\pm$ 0.0009 & 0.1206 $\pm$ 0.0009 \\
AFT&0.0470 $\pm$ 0.0006 & 0.0827 $\pm$ 0.0009 & 0.1107 $\pm$ 0.0009 \\
RSF&0.0447 $\pm$ 0.0006 & 0.0802 $\pm$ 0.0008 & 0.1095 $\pm$ 0.0010 \\
FSN&0.0462 $\pm$ 0.0006 & 0.0800 $\pm$ 0.0008 & 0.1075 $\pm$ 0.0009 \\
DHT&0.0450 $\pm$ 0.0006 & 0.0788 $\pm$ 0.0008 & 0.1074 $\pm$ 0.0010 \\
DSM&0.0451 $\pm$ 0.0006 & 0.0797 $\pm$ 0.0008 & 0.1073 $\pm$ 0.0009 \\ \midrule
\rowcolor{Gray}DCM&0.0450 $\pm$ 0.0006 & 0.0785 $\pm$ 0.0008 & 0.1064 $\pm$ 0.0010 \\ \bottomrule
\end{tabular}}

    \caption{\small Results for various performance metrics on SEER (entire population) along with bootstrapped standard errors.}
    \label{tab:res_seer}
\end{table*}

\begin{table*}[!t]
    \centering
    
    \vspace{1em}
$C^{\text{td}}$($t$) ($\uparrow$)\\
\resizebox{0.685\textwidth}{!}{\begin{tabular}{l|c|c|c}
    \toprule \midrule
     \multirow{2}{*}{ \textbf{Model}} & \multicolumn{3}{|c|}{\textbf{Quantiles}}  \\ \cline{2-4} 
& {$t=25$th} & {$t=50$th} & {$t=75$th}   \\ \hline
CPH&0.8804 $\pm$ 0.0043 & 0.8405 $\pm$ 0.0039 & 0.8121 $\pm$ 0.0037 \\
AFT&0.8865 $\pm$ 0.0042 & 0.8466 $\pm$ 0.0036 & 0.8204 $\pm$ 0.0035 \\
RSF&0.8797 $\pm$ 0.0048 & 0.8379 $\pm$ 0.0038 & 0.8105 $\pm$ 0.0035 \\
FSN&0.8870 $\pm$ 0.0043 & 0.8490 $\pm$ 0.0038 & 0.8248 $\pm$ 0.0036 \\
DHT&0.8920 $\pm$ 0.0039 & 0.8540 $\pm$ 0.0038 & 0.8255 $\pm$ 0.0037 \\
DSM&0.8908 $\pm$ 0.0038 & 0.8506 $\pm$ 0.0038 & 0.8243 $\pm$ 0.0036 \\ \midrule
\rowcolor{Gray}DCM&0.8933 $\pm$ 0.0037 & 0.8558 $\pm$ 0.0036 & 0.8296 $\pm$ 0.0034\\
\bottomrule
\end{tabular}}
    
    \vspace{1em}

    \textsc{AUC($t$)} ($\uparrow$)\\
\resizebox{0.685\textwidth}{!}{\begin{tabular}{l|c|c|c}
    \toprule \midrule
     \multirow{2}{*}{ \textbf{Model}} & \multicolumn{3}{|c|}{\textbf{Quantiles}}  \\ \cline{2-4} 
& {$t=25$th} & {$t=50$th} & {$t=75$th}   \\ \hline
CPH&0.8888 $\pm$ 0.0043 & 0.8604 $\pm$ 0.0042 & 0.8398 $\pm$ 0.0042 \\
AFT&0.8952 $\pm$ 0.0042 & 0.8676 $\pm$ 0.0039 & 0.8491 $\pm$ 0.0040 \\
RSF&0.8867 $\pm$ 0.0048 & 0.8571 $\pm$ 0.0041 & 0.8373 $\pm$ 0.0039 \\
FSN&0.8963 $\pm$ 0.0043 & 0.8702 $\pm$ 0.0040 & 0.8538 $\pm$ 0.0041 \\
DHT&0.9002 $\pm$ 0.0039 & 0.8754 $\pm$ 0.0041 & 0.8540 $\pm$ 0.0041 \\
DSM&0.9033 $\pm$ 0.0036 & 0.8770 $\pm$ 0.0039 & 0.8591 $\pm$ 0.0037 \\ \midrule
\rowcolor{Gray}DCM&0.9020 $\pm$ 0.0037 & 0.8775 $\pm$ 0.0038 & 0.8595 $\pm$ 0.0038\\ \midrule \bottomrule
\end{tabular}}

\vspace{1em}

  \textsc{ECE($t$)} ($\downarrow$)\\
\resizebox{0.685\textwidth}{!}{\begin{tabular}{l|c|c|c}
    \toprule \midrule
     \multirow{2}{*}{ \textbf{Model}} & \multicolumn{3}{|c|}{\textbf{Quantiles}}  \\ \cline{2-4} 
& {$t=25$th} & {$t=50$th} & {$t=75$th}   \\ \hline
CPH&0.0399 $\pm$ 0.0018 & 0.0642 $\pm$ 0.0021 & 0.0764 $\pm$ 0.0028 \\
AFT&0.0173 $\pm$ 0.0016 & 0.0271 $\pm$ 0.0022 & 0.0278 $\pm$ 0.0029 \\
RSF&0.0112 $\pm$ 0.0016 & 0.0219 $\pm$ 0.0023 & 0.0270 $\pm$ 0.0029 \\
FSN&0.0152 $\pm$ 0.0016 & 0.0198 $\pm$ 0.0025 & 0.0196 $\pm$ 0.0029 \\
DHT&0.0107 $\pm$ 0.0015 & 0.0134 $\pm$ 0.0020 & 0.0170 $\pm$ 0.0024 \\
DSM&0.0125 $\pm$ 0.0016 & 0.0292 $\pm$ 0.0023 & 0.0311 $\pm$ 0.0031 \\ \midrule
\rowcolor{Gray}DCM&0.0105 $\pm$ 0.0016 & 0.0145 $\pm$ 0.0024 & 0.0169 $\pm$ 0.0024 
\\ \bottomrule
\end{tabular}}

\vspace{1em}
BS($t$) ($\downarrow$)\\
\resizebox{0.685\textwidth}{!}{\begin{tabular}{l|c|c|c}
    \toprule \midrule
     \multirow{2}{*}{ \textbf{Model}} & \multicolumn{3}{|c|}{\textbf{Quantiles}}  \\ \cline{2-4} 
& {$t=25$th} & {$t=50$th} & {$t=75$th}   \\ \hline
CPH&0.0563 $\pm$ 0.0014 & 0.0989 $\pm$ 0.0019 & 0.1285 $\pm$ 0.0020 \\
AFT&0.0522 $\pm$ 0.0013 & 0.0907 $\pm$ 0.0019 & 0.1168 $\pm$ 0.0021 \\
RSF&0.0508 $\pm$ 0.0014 & 0.0899 $\pm$ 0.0018 & 0.1190 $\pm$ 0.0021 \\
FSN&0.0515 $\pm$ 0.0013 & 0.0877 $\pm$ 0.0017 & 0.1133 $\pm$ 0.0020 \\
DHT&0.0509 $\pm$ 0.0012 & 0.0861 $\pm$ 0.0017 & 0.1135 $\pm$ 0.0021 \\
DSM&0.0509 $\pm$ 0.0013 & 0.0882 $\pm$ 0.0018 & 0.1140 $\pm$ 0.0020 \\ \midrule
\rowcolor{Gray}DCM&0.0508 $\pm$ 0.0012 & 0.0862 $\pm$ 0.0017 & 0.1127 $\pm$ 0.0020 \\ \bottomrule
\end{tabular}}

    \caption{\small Results for various performance metrics on SEER (minority) along with bootstrapped standard errors.}
    \label{tab:res_seer_min}
\end{table*}

Tables \ref{tab:res_seer} and \ref{tab:res_seer_min} and present the $C^{\text{td}}$, AUC, ECE and Brier Score for the Entire Population and Minority Demographic on the SEER dataset, respectively.

\newpage

\begin{figure*}[!t]
    \centering
    { $ \qquad\qquad\qquad\qquad\qquad\qquad C^{\text{td}}$ \hfill  $T=25^{\textbf{th}}$ \hfill    $\text{ECE}\quad\qquad\qquad\qquad\qquad\qquad$  }\\
    \includegraphics[width=0.495\textwidth]{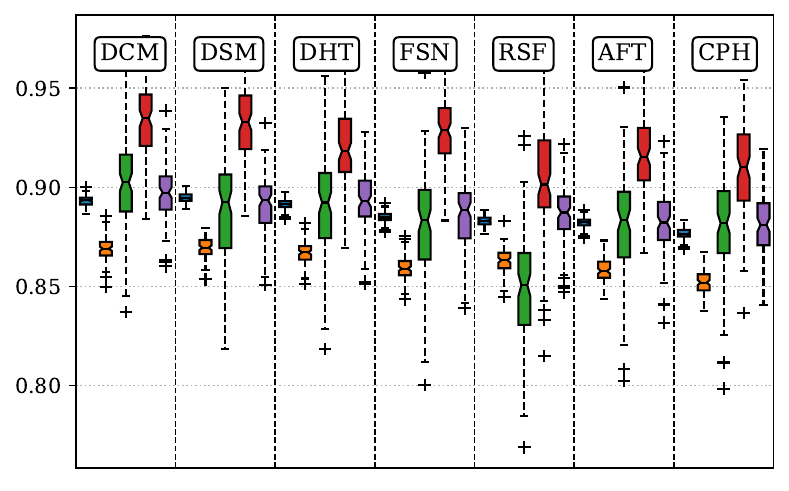} \hfill
    \includegraphics[width=0.495\textwidth]{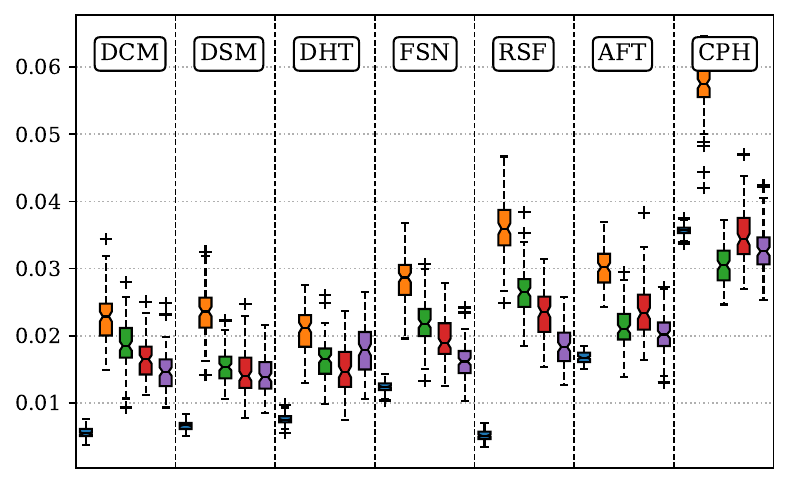}\\
    { $ \qquad\qquad\qquad\qquad\qquad\qquad C^{\text{td}}$ \hfill  $T=50^{\textbf{th}}$ \hfill    $\text{ECE}\quad\qquad\qquad\qquad\qquad\qquad$  }\\
    \includegraphics[width=0.495\textwidth]{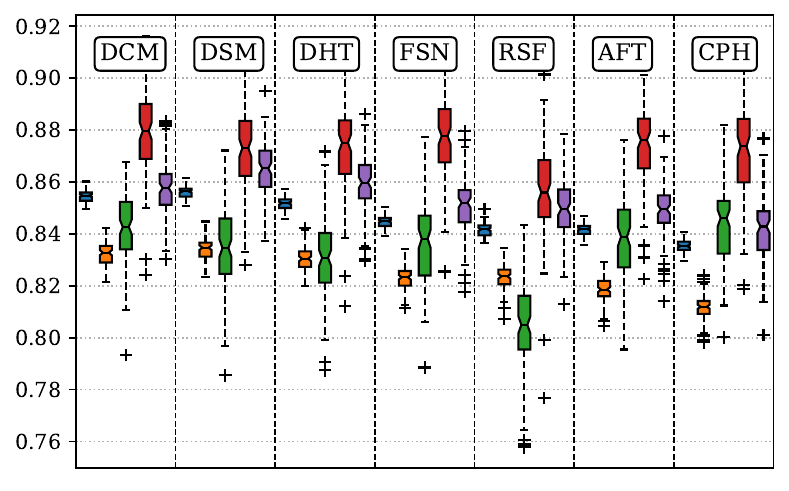} \hfill
    \includegraphics[width=0.495\textwidth]{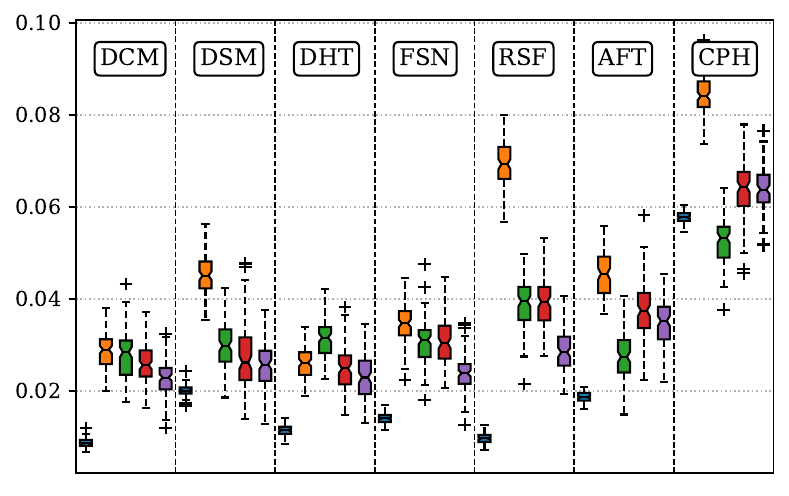}\\
    { $ \qquad\qquad\qquad\qquad\qquad\qquad C^{\text{td}}$ \hfill  $T=75^{\textbf{th}}$ \hfill    $\text{ECE}\quad\qquad\qquad\qquad\qquad\qquad$  }\\
    \includegraphics[width=0.495\textwidth]{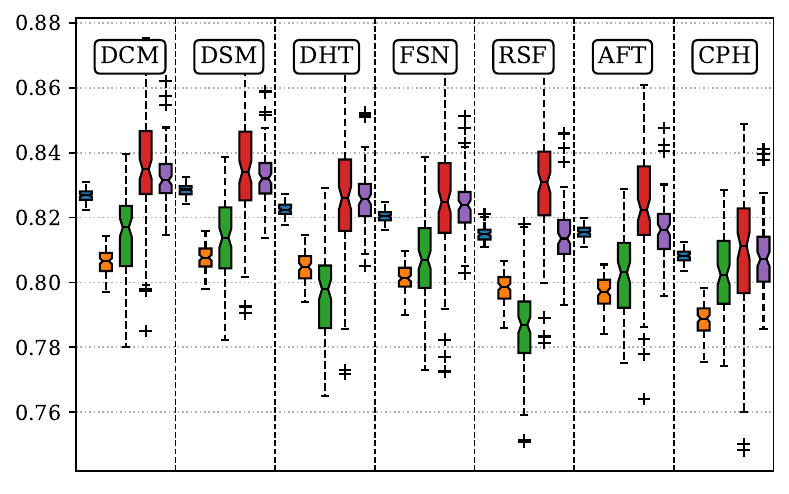} \hfill
    \includegraphics[width=0.495\textwidth]{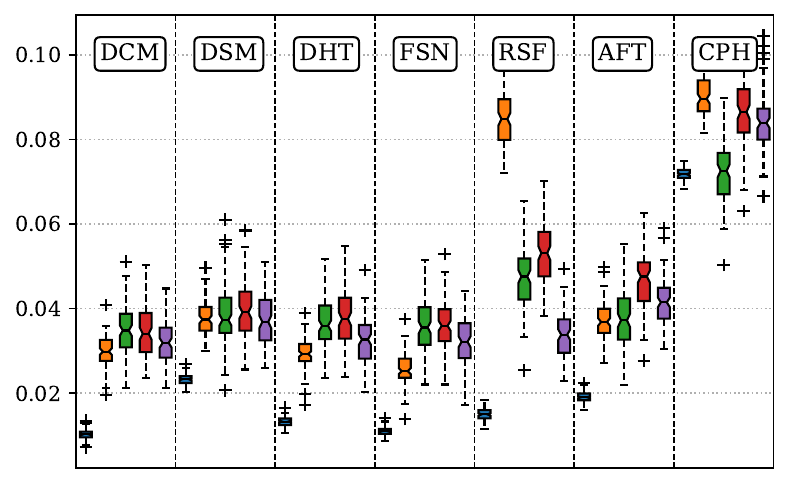}\\
    \includegraphics[width=.85\textwidth]{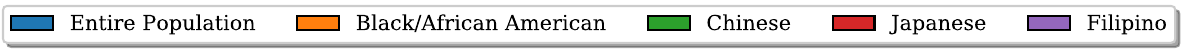}\\
    \caption{$C^{\text{td}}$ (higher means better discrimination) and ECE (lower means better calibration) of proposed approach versus baselines at different quantiles of event times for the minority demographics. The rows represents different quantiles at which we evaluate the individual metrics. (Minorities in the dataset are denoted by different colors in the legend)}
    \label{fig:seer_all_minorities}
    
\end{figure*}

SEER has multiple minority classes. In Figure \ref{fig:seer_all_minorities} we break results down by the top four largest minorities in the subset of SEER we are working with, `Black/African American', `Chinese', `Japanese' and `Filipino'.

\newpage
\subsection{Unawareness to Group Membership}
\label{apx:unawareness}

\begin{table*}[!t]
\centering
\begin{minipage}{0.5\textwidth}
\centering
$C^{\text{td}}$($t$) ($\uparrow$)\\
\resizebox{0.9\textwidth}{!}{\begin{tabular}{l|c|c|c}
    \toprule \midrule
     \multirow{2}{*}{ \textbf{Demographic}} & \multicolumn{3}{|c}{\textbf{Quantiles}}  \\ \cline{2-4} 
& {$t=25$th} & {$t=50$th} & {$t=75$th}   \\ \hline
Entire Population&- 1.05 \% & -1.34 \% & -1.36 \% \\
Minority Group&- 1.29 \% & -1.48 \% & -1.73 \% \\ \midrule \bottomrule
\end{tabular}}
\end{minipage}%
\begin{minipage}{0.5\textwidth}
\centering
AUC($t$) ($\uparrow$)\\
\resizebox{0.9\textwidth}{!}{\begin{tabular}{l|c|c|c}
    \toprule \midrule
     \multirow{2}{*}{ \textbf{Demographic}} & \multicolumn{3}{c}{\textbf{Quantiles}}  \\ \cline{2-4} 
& {$t=25$th} & {$t=50$th} & {$t=75$th}   \\ \hline
Entire Population&- 1.07 \% & - 1.42 \% & - 1.53 \% \\
Minority Group&- 1.36 \% & - 1.59 \% & - 1.95 \% \\ \midrule \bottomrule
\end{tabular}}
\end{minipage}

\vspace{2em}

\begin{minipage}{0.5\textwidth}
\centering
ECE($t$) ($\downarrow$)\\
\resizebox{0.9\textwidth}{!}{\begin{tabular}{l|c|c|c}
    \toprule \midrule
     \multirow{2}{*}{ \textbf{Demographic}} & \multicolumn{3}{c}{\textbf{Quantiles}}  \\ \cline{2-4} 
& {$t=25$th} & {$t=50$th} & {$t=75$th}   \\ \hline
Entire Population& + 17.30 \% & + 13.12 \% & + 10.74 \% \\
Minority Group& - 0.80 \% & + 19.85 \% & + 20.99 \% \\ \midrule \bottomrule
\end{tabular}}
\end{minipage}%
\begin{minipage}{0.5\textwidth}
\centering
BS($t$) ($\downarrow$)\\
\resizebox{0.9\textwidth}{!}{\begin{tabular}{l|c|c|c}
    \toprule \midrule
     \multirow{2}{*}{ \textbf{Demographic}} & \multicolumn{3}{c}{\textbf{Quantiles}}  \\ \cline{2-4} 
& {$t=25$th} & {$t=50$th} & {$t=75$th}   \\ \hline
Entire Population& + 1.89 \% & + 2.80 \% & + 3.13 \% \\
Minority Group& + 2.07 \% & + 4.53 \% & + 4.70 \% \\ \midrule \bottomrule
\end{tabular}}
\end{minipage}

\caption{Relative change in performance of Deep Cox Mixtures on the SEER dataset for the Entire Population and the Minority Demographic when unaware of the protected group membership. Overall, the performance in terms of both Discrimination and Calibration drops when DCM is made unaware of the protected groups. The relative deterioration in performance is worse for the minority demographic, suggesting unawareness to protected attribute being harmful in terms of the above performance metrics. }
    \label{tab:res_unawareness}
\end{table*}

In Table \ref{tab:res_unawareness} we attempt to see how unawareness to the demographic affects the performance of Deep Cox Mixtures in terms of both, Calibration and Discrimination.

\subsection{Dynamics of the Proposed MCMC EM Algorithm}

\begin{figure}[!htbp]
    \centering
    \includegraphics[width=0.65\textwidth]{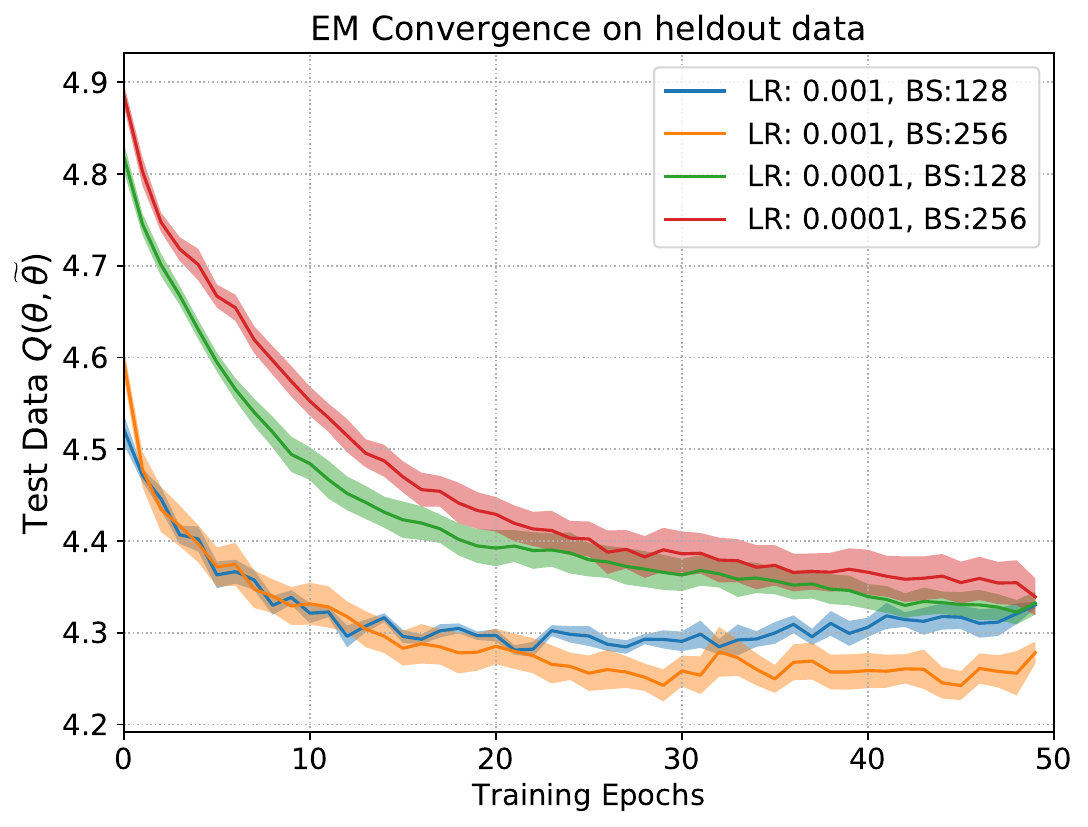}

    \caption{The estimated $Q(\theta, \widetilde{\theta})$ on a heldout set from the SUPPORT dataset for different hyper-parameters (LR: Learning Rate, BS: Batch size).   }
    \label{fig:dcmem}
\end{figure}

Figure \ref{fig:dcmem} presents the estimated $Q(\theta, \widetilde{\theta})$ funciton for healdout dataset for the SUPPORT dataset. Empirically our proposed monte carlo EM monotonically decreases the $Q(\theta, \widetilde{\theta})$ suggesting good learning dynamics.

\bibliographystylesup{apalike}
\bibliographysup{ref_supp}
\end{document}